\documentclass[11pt, a4paper, logo, copyright, nonumbering]{deepmind}
\usepackage[authoryear, sort&compress, round]{natbib}
\usepackage{dblfloatfix}
\usepackage{ulem}
\usepackage{caption}
\usepackage{subcaption}
\usepackage{listings}
\usepackage{longtable}
\usepackage{dramatist}
\usepackage{xspace}
\usepackage{pifont}% http://ctan.org/pkg/pifont
\usepackage{multirow}
\usepackage{tcolorbox}
\usepackage{xltabular}
\usepackage{longtable}
\usepackage{hyperref}
\usepackage{amsfonts}
\usepackage{amsmath}
\usepackage{amssymb}
\usepackage{lineno}

\usepackage[bottom]{footmisc}
\makeatletter
\def\@BTrule[#1]{%
  \ifx\longtable\undefined
    \let\@BTswitch\@BTnormal
  \else\ifx\hline\LT@hline
    \nobreak
    \let\@BTswitch\@BLTrule
  \else
     \let\@BTswitch\@BTnormal
  \fi\fi
  \global\@thisrulewidth=#1\relax
  \ifnum\@thisruleclass=\tw@\vskip\@aboverulesep\else
  \ifnum\@lastruleclass=\z@\vskip\@aboverulesep\else
  \ifnum\@lastruleclass=\@ne\vskip\doublerulesep\fi\fi\fi
  \@BTswitch}
\makeatother

\addto\extrasenglish{
}

 {\begin{list}{}%
         {\setlength{\leftmargin}{#1}}%
         \item[]%
 }
 {\end{list}}
 
\newcommand{\massivetext}{\textit{MassiveText}\xspace}
\newcommand{\massiveweb}{\textit{MassiveWeb}\xspace}

\newcommand{\gopher}{\textit{Gopher}\xspace}
\newcommand{\nummodels}{400 }
\newcommand{\Gopher}{\textit{Gopher}\xspace}
\newcommand{\chinchilla}{\textit{Chinchilla}\xspace}

\newcommand{\mtnlg}{MT-NLG 530B\xspace}
\newcommand{\bigbench}{BIG-bench\xspace}

\correspondingauthor{email@email}

\reportnumber{001} % Leave blank if n/a

\renewcommand{\today}{}

\title{\centering Training Compute-Optimal Large Language Models}

\correspondingauthor{ \{jordanhoffmann|sborgeaud|amensch|sifre\}@deepmind.com}

\author[$\star$]{Jordan~Hoffmann}
\author[$\star$]{Sebastian~Borgeaud}
\author[$\star$]{Arthur~Mensch}

\author[  \hspace{-.8ex}]{Elena~Buchatskaya}
\author[  \hspace{-.8ex}]{Trevor~Cai}
\author[  \hspace{-.8ex}]{Eliza~Rutherford}
\author[  \hspace{-.8ex}]{Diego~de~Las~Casas}
\author[  \hspace{-.8ex}]{Lisa~Anne~Hendricks}
\author[  \hspace{-.8ex}]{Johannes~Welbl}
\author[  \hspace{-.8ex}]{Aidan~Clark}
\author[  \hspace{-.8ex}]{Tom~Hennigan}
\author[  \hspace{-.8ex}]{Eric~Noland}
\author[  \hspace{-.8ex}]{Katie~Millican}
\author[  \hspace{-.8ex}]{George~van~den~Driessche}
\author[  \hspace{-.8ex}]{Bogdan~Damoc}
\author[  \hspace{-.8ex}]{Aurelia~Guy}
\author[  \hspace{-.8ex}]{Simon~Osindero}
\author[  \hspace{-.8ex}]{Karen~Simonyan}
\author[  \hspace{-.8ex}]{Erich~Elsen}
\author[  \hspace{-.8ex}]{Jack~W.~Rae}
\author[  \hspace{-.8ex}]{Oriol~Vinyals}
\author[$\star$]{Laurent~Sifre}

\affil[$\star$]{Equal contributions}

\usepackage[titles]{tocloft}
\newcommand{\EE}{\mathbb{E}}

\newcommand{\Dd}{\mathcal{D}}

\newcommand{\Ff}{\mathcal{F}}

\newcommand{\Hh}{\mathcal{H}}

\newcommand{\Xx}{\mathcal{X}}
\newcommand{\Yy}{\mathcal{Y}}

\renewcommand{\phi}{\varphi}

\renewcommand{\leq}{\leqslant}

\renewcommand{\epsilon}{\varepsilon}
\renewcommand{\imath}{\mathrm{i}}

\DeclareMathOperator*{\argmin}{argmin}

\newlength{\restsubwidth}
\newlength{\restsubheight}
\newlength{\restsubmoreheight}
\newcommand{\rest}[2]{%
        \settowidth{\restsubwidth}{\ensuremath{#2}}
        \settoheight{\restsubheight}{\ensuremath{{}_{#2}}}
        \ensuremath{{#1\hskip 0.5pt}_{\vrule\kern2pt\parbox[b][%
        4pt][b]{\the\restsubwidth}{%
                        \ensuremath{{}_{#2}}}}}
        }

\begin{abstract}
We investigate the optimal model size and number of tokens for training a transformer language model under a given compute budget.
We find that current large language models are significantly undertrained, a consequence of the recent focus on scaling language models whilst keeping the amount of training data constant.
By training over \nummodels language models ranging from 70 million to over 16 billion parameters on 5 to 500 billion tokens, we find that for compute-optimal training, the model size and the number of training tokens should be scaled equally: for every doubling of model size the number of training tokens should also be doubled.
We test this hypothesis by training a predicted compute-optimal model, \chinchilla, that uses the same compute budget as \gopher but with 70B parameters and 4$\times$ more more data.
\chinchilla uniformly and significantly outperforms \Gopher (280B), GPT-3 (175B), Jurassic-1 (178B), and Megatron-Turing NLG (530B) on a large range of downstream evaluation tasks.
This also means that \chinchilla uses substantially less compute for fine-tuning and inference, greatly facilitating downstream usage. 
As a highlight, \chinchilla reaches a state-of-the-art average accuracy of 67.5\% on the MMLU benchmark, greater than a 7\% improvement over \gopher. 
\end{abstract}

\begin{document}

\maketitle

\section{Introduction}
Recently a series of \textit{Large Language Models} (LLMs) have been introduced \citep{gpt3, jurassic, rae2021gopher, nlg530b, thoppilan2022lamda}, with the largest dense language models now having over 500 billion parameters.
These large autoregressive transformers \citep{vaswani2017attention} have demonstrated impressive performance on many tasks using a variety of evaluation protocols such as zero-shot, few-shot, and fine-tuning.

The compute and energy cost for training large language models is substantial \citep{rae2021gopher, thoppilan2022lamda} and rises with increasing model size.
In practice, the allocated training compute budget is often known in advance: how many accelerators are available and for how long we want to use them.
Since it is typically only feasible to train these large models once, accurately estimating the best model hyperparameters for a given compute budget is critical \citep{tay2021scale}.

\citet{kaplan2020scaling} showed that there is a power law relationship between the number of parameters in an autoregressive language model (LM) and its performance.
As a result, the field has been training larger and larger models, expecting performance improvements. 
One notable conclusion in \citet{kaplan2020scaling} is that large models should not be trained to their lowest possible loss to be compute optimal. 
Whilst we reach the same conclusion, we estimate that large models should be trained for many more training tokens than recommended by the authors.
Specifically, given a $10\times$ increase computational budget, they suggests that the size of the model should increase $5.5\times$ while the number of training tokens should only increase 1.8$\times$.
Instead, we find that model size and the number of training tokens should be scaled in equal proportions.

Following \citet{kaplan2020scaling} and the training setup of GPT-3 \citep{gpt3}, many of the recently trained large models have been trained for approximately 300 billion tokens (\autoref{tab:llms}), in line with the approach of predominantly increasing model size when increasing compute.

\begin{figure*}[t]
    \centering
    \includegraphics[width=.9\textwidth, trim=-5cm 0 0 0]{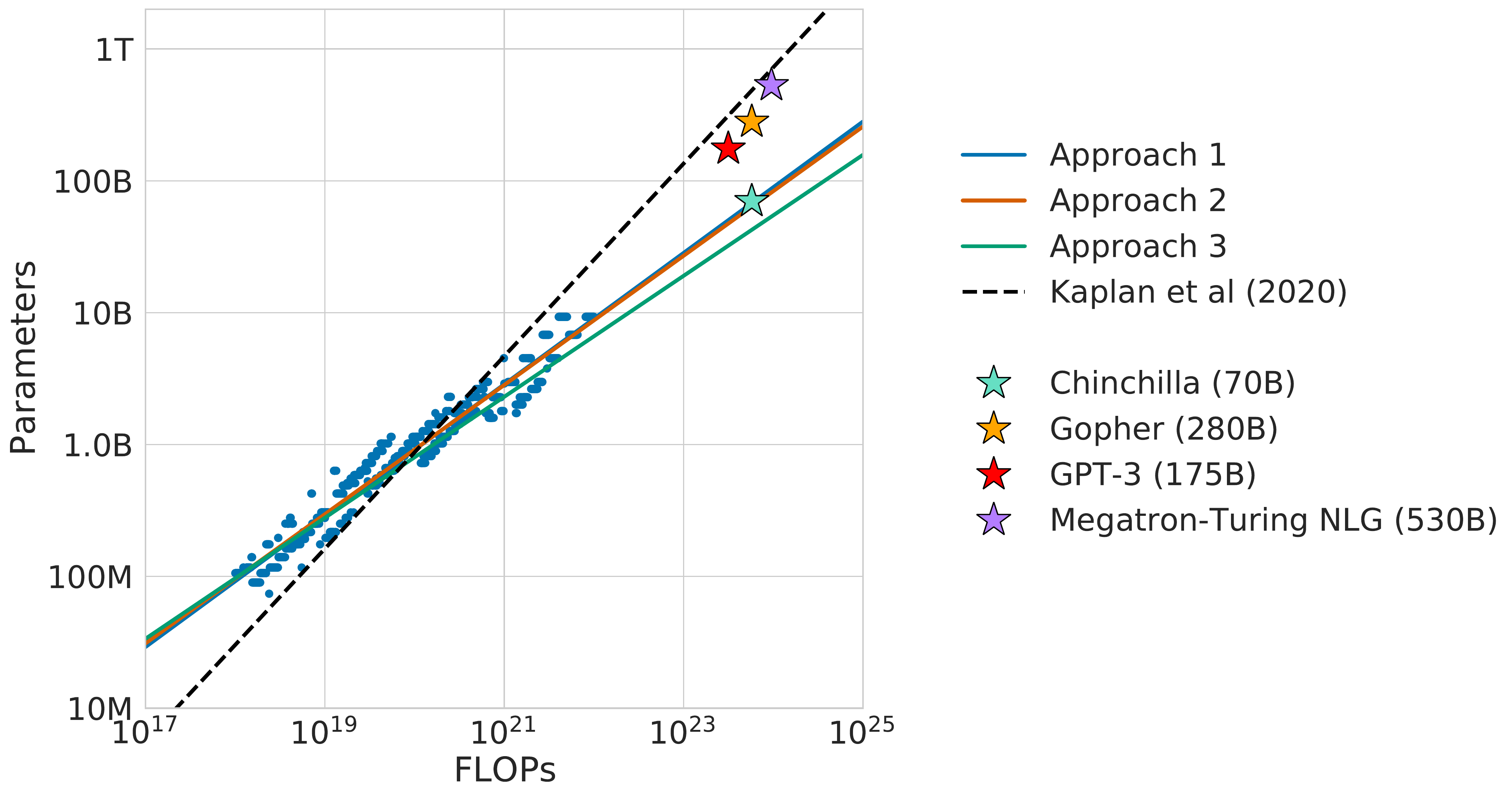}
    \caption{\textbf{Overlaid predictions.}
    We overlay the predictions from our three different approaches, along with projections from \citet{kaplan2020scaling}. We find that all three methods predict that current large models should be substantially smaller and therefore trained much longer than is currently done.
    In \autoref{fig:token_flop}, we show the results with the predicted optimal tokens plotted against the optimal number of parameters for fixed FLOP budgets.
    \textbf{\chinchilla outperforms \gopher and the other large models (see \autoref{sec:model_analysis}).}
    }
    \label{fig:combined_predictions}
\end{figure*}

\begin{table*}[t]
    \caption{\textbf{Current LLMs}.
    We show five of the current largest dense transformer models, their size, and the number of training tokens. Other than LaMDA \citep{thoppilan2022lamda}, most models are trained for approximately 300 billion tokens. 
    We introduce \chinchilla, a substantially smaller model, trained for much longer than 300B tokens.
    }
    \label{tab:llms}
\centering
\begin{tabular}{l r c}
\toprule
Model & Size ($\#$ Parameters) & Training Tokens \\
\midrule
LaMDA \citep{thoppilan2022lamda} & 137 Billion &168 Billion \\
GPT-3 \citep{gpt3} & 175 Billion & 300 Billion \\
Jurassic \citep{jurassic} & 178 Billion & 300 Billion \\
\gopher \citep{rae2021gopher} & 280 Billion & 300 Billion \\
\mtnlg \citep{nlg530b} & 530 Billion & 270 Billion\\
\midrule
\chinchilla & 70 Billion & 1.4 Trillion\\
\bottomrule
\end{tabular}
\end{table*}

In this work, we revisit the question:
\textit{Given a fixed FLOPs budget,\footnote{For example, knowing the number of accelerators and a target training duration.} how should one trade-off model size and the number of training tokens?}
To answer this question, we model the final pre-training loss\footnote{
For simplicity, we perform our analysis on the smoothed training loss which is an unbiased estimate of the test loss, as we are in the infinite data regime (the number of training tokens is less than the number of tokens in the entire corpus).}
$L(N, D)$ as a function of the number of model parameters~$N$, and the number of training tokens,~$D$.
Since the computational budget $C$ is a deterministic function $\text{FLOPs}(N,D)$ of the number of seen training tokens and model parameters, we are interested in minimizing $L$ under the constraint $\text{FLOPs}(N, D) = C$:
\begin{equation}\label{eq:model}
    N_{opt}(C), D_{opt}(C) = \argmin_{N, D \text{ s.t. } \text{FLOPs}(N, D) = C} L(N, D).
\end{equation}
The functions $N_{opt}(C)$, and $D_{opt}(C)$ describe the optimal allocation of a computational budget $C$. 
We empirically estimate these functions based on the losses of over \nummodels models, ranging from under $70$M to over $16$B parameters, and trained on $5$B to over $400$B tokens -- with each model configuration trained for several different training horizons.
Our approach leads to considerably different results than that of \citet{kaplan2020scaling}.
We highlight our results in \autoref{fig:combined_predictions} and how our approaches differ in \autoref{sec:related_work}.

Based on our estimated compute-optimal frontier, we predict that for the compute budget used to train \gopher, an optimal model should be 4 times smaller, while being training on 4 times more tokens.
We verify this by training a more \textit{compute-optimal} 70B model, called \chinchilla, on 1.4 trillion tokens. 
Not only does \chinchilla outperform its much larger counterpart, \gopher, but its reduced model size reduces inference cost considerably and greatly facilitates downstream uses on smaller hardware.
The energy cost of a large language model is amortized through its usage for inference an fine-tuning.
The benefits of a more optimally trained smaller model, therefore, extend beyond the immediate benefits of its improved performance.  

\section{Related Work}
\label{sec:related_work}
\paragraph{Large language models.}
A variety of large language models have been introduced in the last few years.
These include both dense transformer models \citep{gpt3, jurassic, nlg530b, rae2021gopher, thoppilan2022lamda} and mixture-of-expert (MoE) models \citep{du2021glam, fedus2021switch, zoph2022designing}. 
The largest dense transformers have passed 500 billion parameters \citep{nlg530b}.
The drive to train larger and larger models is clear---so far increasing the size of language models has been responsible for improving the state-of-the-art in many language modelling tasks. 
Nonetheless, large language models face several challenges, including their overwhelming computational requirements (the cost of training and inference increase with model size) \citep{rae2021gopher, thoppilan2022lamda} and the need for acquiring more high-quality training data.
In fact, in this work we find that larger, high quality datasets will play a key role in any further scaling of language models.

\paragraph{Modelling the scaling behavior.}
Understanding the scaling behaviour of language models and their transfer properties has been important in the development of recent large models \citep{kaplan2020scaling,  hernandez2021scaling}. 
\citet{kaplan2020scaling} first showed a predictable relationship between model size and loss over many orders of magnitude.
The authors investigate the question of choosing the optimal model size to train for a given compute budget. Similar to us, they address this question by training various models.
Our work differs from \citet{kaplan2020scaling} in several important ways.
First, the authors use a fixed number of training tokens and learning rate schedule for all models; this prevents them from modelling the impact of these hyperparameters on the loss.
In contrast, we find that setting the learning rate schedule to approximately match the number of training tokens results in the best final loss regardless of model size---see \autoref{fig:cosine}.
For a fixed learning rate cosine schedule to 130B tokens, the intermediate loss estimates (for $D' << 130$B) are therefore overestimates of the loss of a model trained with a schedule length matching $D'$.
Using these intermediate losses results in underestimating the effectiveness of training models on less data than 130B tokens, and eventually contributes to the conclusion that model size should increase faster than training data size as compute budget increases.
In contrast, our analysis predicts that both quantities should scale at roughly the same rate.
Secondly, we include models with up to 16B parameters, as we observe that there is slight curvature in the FLOP-loss frontier (see \autoref{app:curvature})---in fact, the majority of the models used in our analysis have more than 500 million parameters, in contrast the majority of runs in \citet{kaplan2020scaling} are significantly smaller---many being less than 100M parameters.

Recently, \citet{clark2022unified} specifically looked in to the scaling properties of Mixture of Expert language models, showing that the scaling with number of experts diminishes as the model size increases---their approach models the loss as a function of two variables: the model size and the number of experts.
However, the analysis is done with a fixed number of training tokens, as in \citet{kaplan2020scaling}, potentially underestimating the improvements of branching.

\paragraph{Estimating hyperparameters for large models.}
The model size and the number of training tokens are not the only two parameters to chose when selecting a language model and a procedure to train it.
Other important factors include learning rate, learning rate schedule, batch size, optimiser, and width-to-depth ratio.
In this work, we focus on model size and the number of training steps, and we rely on existing work and provided experimental heuristics to determine the other necessary hyperparameters.
\citet{yang2021tuning} investigates how to choose a variety of these parameters for training an autoregressive transformer, including the learning rate and batch size.
\citet{mccandlish2018empirical} finds only a weak dependence between optimal batch size and model size. \citet{shallue2018measuring, NEURIPS2019_e0eacd98} suggest that using larger batch-sizes than those we use is possible.
\citet{levine2020depth} investigates the optimal depth-to-width ratio for a variety of standard model sizes. We use slightly less deep models than proposed as this translates to better wall-clock performance on our hardware. 

\paragraph{Improved model architectures.}
Recently, various promising alternatives to traditional dense transformers have been proposed.
For example, through the use of conditional computation large MoE models like the 1.7 trillion parameter Switch transformer \citep{fedus2021switch}, the 1.2 Trillion parameter GLaM model \citep{du2021glam}, and others \citep{artetxe2021efficient, zoph2022designing} are able to provide a large effective model size despite using relatively fewer training and inference FLOPs.
However, for very large models the computational benefits of routed models seems to diminish \citep{clark2022unified}.
An orthogonal approach to improving language models is to augment transformers with explicit retrieval mechanisms, as done by \citet{borgeaud2021retrieval, guu2020realm, lewisretrieval2020}. This approach effectively increases the number of data tokens seen during training (by a factor of $\sim10$ in \citet{borgeaud2021retrieval}).
This suggests that the performance of language models may be more dependant on the size of the training data than previously thought.

\section{Estimating the optimal parameter/training tokens allocation}
\label{sec:method}
We present three different approaches to answer the question driving our research: 
\textit{Given a fixed FLOPs budget, how should one trade-off model size and the number of training tokens?} 
In all three cases we start by training a range of models varying both model size and the number of training tokens and use the resulting training curves to fit an empirical estimator of how they should scale. 
We assume a power-law relationship between compute and model size as done in \citet{clark2022unified, kaplan2020scaling}, though future work may want to include potential curvature in this relationship for large model sizes.
The resulting predictions are similar for all three methods and suggest that parameter count and number of training tokens should be increased equally with more compute\footnote{We compute FLOPs as described in \autoref{sec:flops}.}---with proportions reported in \autoref{tab:comparison}.
This is in clear contrast to previous work on this topic and warrants further investigation.

\subsection{Approach 1: Fix model sizes and vary number of training tokens}

In our first approach we vary the number of training steps for a fixed family of models (ranging from 70M to over 10B parameters), training each model for 4 different number of training sequences.
From these runs, we are able to directly extract an estimate of the minimum loss achieved for a given number of training FLOPs. Training details for this approach can be found in \autoref{sec:scaling_details}.

\begin{figure*}[t]
    \centering
    \includegraphics[width=.95\textwidth]{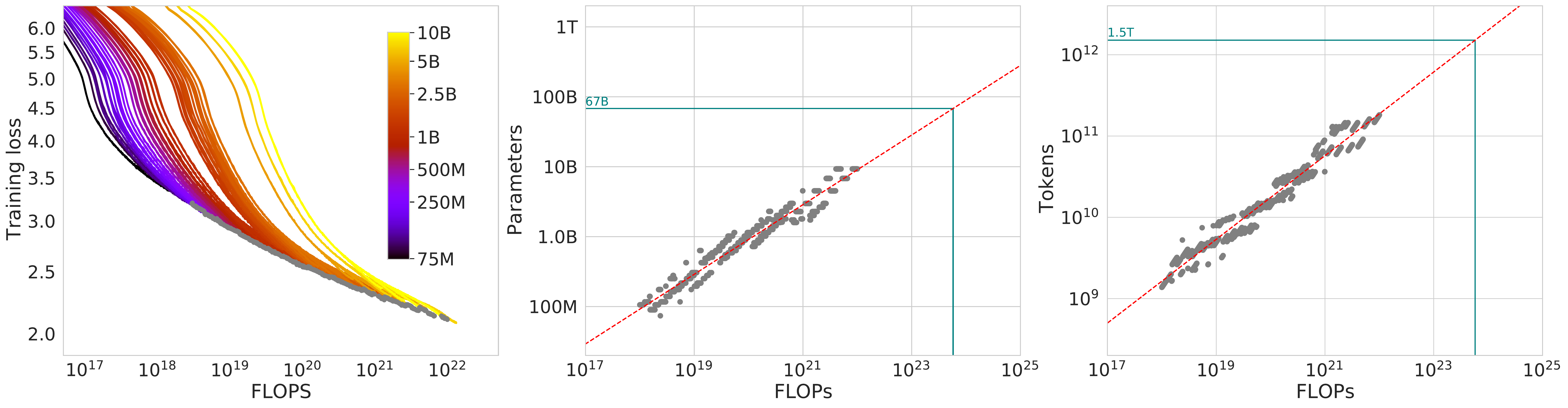}
    \caption{\textbf{Training curve envelope.} On the \textbf{left} we show all of our different runs. We launched a range of model sizes going from 70M to 10B, each for four different cosine cycle lengths. From these curves, we extracted the envelope of minimal loss per FLOP, and we used these points to estimate the optimal model size (\textbf{center}) for a given compute budget and the optimal number of training tokens (\textbf{right}). 
    In green, we show projections of optimal model size and training token count based on the number of FLOPs used to train \gopher ($5.76 \times 10^{23}$).
    }
    \label{fig:approach1}
\end{figure*}

For each parameter count $N$ we train 4 different models, decaying the learning rate by a factor of 10$\times$ over a horizon (measured in number of training tokens) that ranges by a factor of $16 \times$.
Then, for each run, we smooth and then interpolate the training loss curve.
From this, we obtain a continuous mapping from FLOP count to training loss for each run.
Then, for each FLOP count, we determine which run achieves the lowest loss.
Using these interpolants, we obtain a mapping from any FLOP count $C$, to the most efficient choice of model size $N$ and number of training tokens $D$ such that $\text{FLOPs}(N,D) = C$.\footnote{Note that all selected points are within the last 15\% of training. This suggests that when training a model over $D$ tokens, we should pick a cosine cycle length that decays $10 \times$ over approximately $D$ tokens---see further details in \autoref{sec:cosine_cycle}.}
At 1500 logarithmically spaced FLOP values, we find which model size achieves the lowest loss of all models along with the required number of training tokens.
Finally, we fit power laws to estimate the optimal model size and number of training tokens for any given amount of compute (see the center and right panels of \autoref{fig:approach1}), obtaining a relationship $N_{opt} \propto C^a$ and $D_{opt} \propto C^b$.
We find that $a=0.50$ and $b=0.50$---as summarized in \autoref{tab:comparison}.
In \autoref{app:kaplan_comparison}, we show a head-to-head comparison at $10^{21}$ FLOPs, using the model size recommended by our analysis and by the analysis of \citet{kaplan2020scaling}---using the model size we predict has a clear advantage.

\subsection{Approach 2: IsoFLOP profiles}
In our second approach we vary the model size\footnote{In approach 2, model size varies up to 16B as opposed to approach 1 where we only used models up to 10B.} for a fixed set of 9 different training FLOP counts\footnote{The number of training tokens is determined by the model size and training FLOPs.} (ranging from $6 \times10^{18}$ to  $3 \times 10^{21}$ FLOPs), and consider the final training loss for each point\footnote{We set the cosine schedule length to match the number of tokens, which is optimal according to the analysis presented in \autoref{sec:cosine_cycle}.}. in contrast with Approach 1 that considered points $(N, D, L)$ along the entire training runs.
This allows us to directly answer the question: For a given FLOP budget, what is the optimal parameter count?

\begin{figure*}[t]
    \centering
    \includegraphics[width=.95\textwidth]{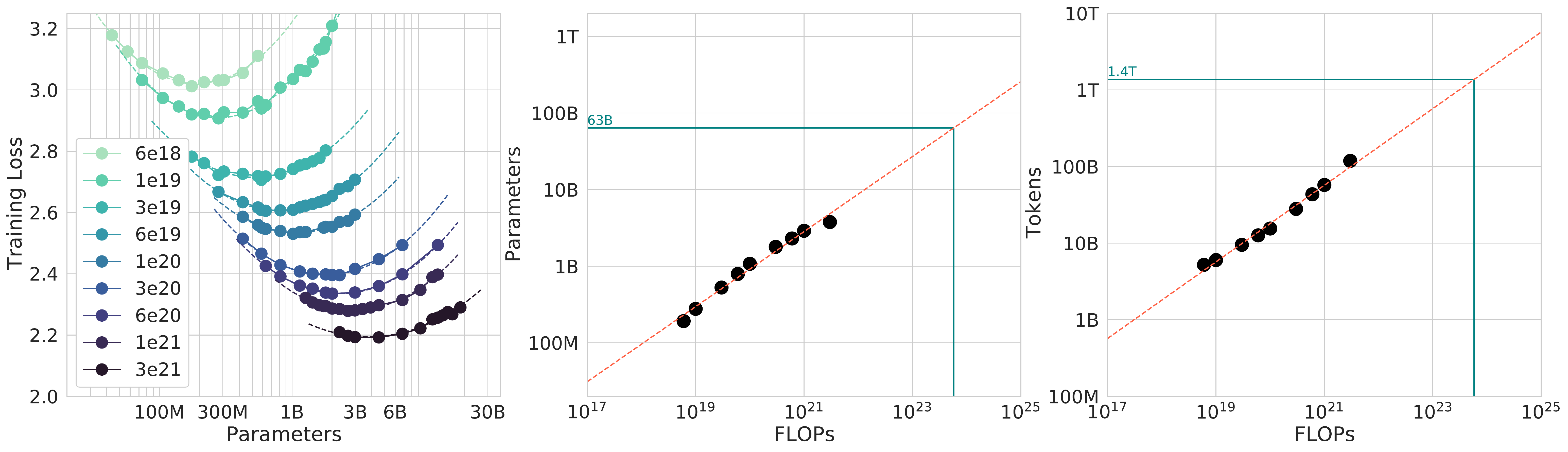}
    \caption{\textbf{IsoFLOP curves.} For various model sizes, we choose the number of training tokens such that the final FLOPs is a constant. The cosine cycle length is set to match the target FLOP count. We find a clear valley in loss, meaning that for a given FLOP budget there is an optimal model to train (\textbf{left}). 
    Using the location of these valleys, we project optimal model size and number of tokens for larger models (\textbf{center} and \textbf{right}).
    In green, we show the estimated number of parameters and tokens for an \textit{optimal} model trained with the compute budget of \gopher.
    }
    \label{fig:isoflop}
\end{figure*}

For each FLOP budget, we plot the final loss (after smoothing) against the parameter count in \autoref{fig:isoflop} (left).
In all cases, we ensure that we have trained a diverse enough set of model sizes to see a clear minimum in the loss. 
We fit a parabola to each IsoFLOPs curve to directly estimate at what model size the minimum loss is achieved (\autoref{fig:isoflop} (left)).
As with the previous approach, we then fit a power law between FLOPs and loss-optimal model size and number of training tokens, shown in \autoref{fig:isoflop} (center, right).
Again, we fit exponents of the form $N_{opt} \propto C^a$ and $D_{opt} \propto C^b$
and we find that $a=0.49$ and $b=0.51$---as summarized in \autoref{tab:comparison}.

\subsection{Approach 3: Fitting a parametric loss function}
Lastly, we model all final losses from experiments in Approach 1 \& 2 as a parametric function of model parameter count and the number of seen tokens.
Following a classical risk decomposition (see \autoref{sec:approach3}), we propose the following functional form
\begin{equation}
    \hat L(N,D) \triangleq E + \frac{A}{N^\alpha} + \frac{B}{D^\beta}.\label{eq:decompose}
\end{equation}
The first term captures the loss for an ideal generative process on the data distribution, and should correspond to the entropy of natural text.
The second term captures the fact that a perfectly trained transformer with $N$ parameters underperforms the ideal generative process.
The final term captures the fact that the transformer is not trained to convergence, as we only make a finite number of optimisation steps, on a sample of the dataset distribution.

\paragraph{Model fitting.} To estimate $(A, B, E, \alpha, \beta)$, we minimize the Huber loss \citep{huber_robust_1964} between the predicted and observed log loss using the L-BFGS algorithm \citep{nocedal_updating_1980}:
\begin{align}
    \min_{A, B, E, \alpha, \beta}\quad &\sum_{\text{Runs }i} \text{Huber}_\delta \Big(\log \hat L(N_i, D_i) - \log L_i\Big) \label{eq:huber}
\end{align}
We account for possible local minima by selecting the best fit from a grid of initialisations.
The Huber loss ($\delta=10^{-3}$) is robust to outliers, which we find important for good predictive performance over held-out data points. \autoref{app:parametric} details the fitting procedure and the loss decomposition.

\begin{figure}
    \centering
    \includegraphics[width=\textwidth]{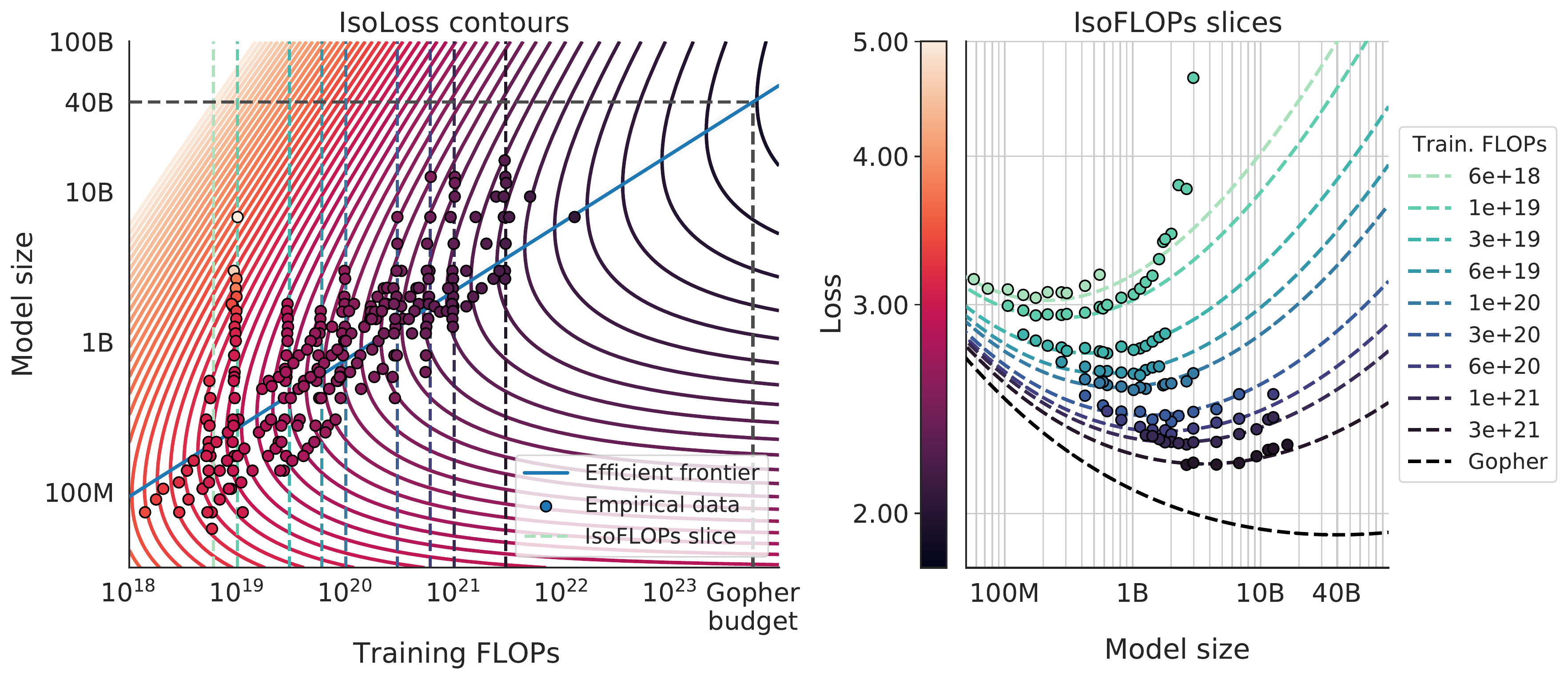}
    \caption{\textbf{Parametric fit.} We fit a parametric modelling of the loss $\hat L(N,D)$ and display contour (\textbf{left}) and isoFLOP slices (\textbf{right}).
    For each isoFLOP slice, we include a corresponding dashed line in the left plot.
    In the left plot, we show the efficient frontier in blue, which is a line in log-log space. 
    Specifically, the curve goes through each iso-loss contour at the point with the fewest FLOPs.
    We project the optimal model size given the \gopher FLOP budget to be 40B parameters.}
    \label{fig:approach_3}
\end{figure}

\paragraph{Efficient frontier.} 
We can approximate the functions $N_{opt}$ and $D_{opt}$ by minimizing the parametric loss $\hat L$ under the constraint
$\text{FLOPs}(N, D) \approx 6 N D$ \citep{kaplan2020scaling}.
The resulting $N_{opt}$ and $D_{opt}$ balance the two terms in Equation~\eqref{eq:huber} that depend on model size and data.
By construction, they have a power-law form:
\begin{equation}
    N_{opt}(C) = G {\left(\frac{C}{6}\right)}^{a}, \quad
    D_{opt}(C) = G^{-1} {\left(\frac{C}{6}\right)}^{b}, \quad \text{ where }\quad
    G = {\left(\frac{\alpha A}{\beta B} \right)}^{\frac{1}{\alpha + \beta}},\quad
    a = \frac{\beta}{\alpha+\beta}, \text{ and } b = \frac{\alpha}{\alpha + \beta}.
\end{equation}
We show contours of the fitted function $\hat L$ in \autoref{fig:approach_3} (left), and the closed-form efficient computational frontier in blue.
From this approach, we find that $a=0.46$ and $b=0.54$---as summarized in \autoref{tab:comparison}.

\subsection{Optimal model scaling}
\label{sec:scaling-results}
We find that the three approaches, despite using different fitting methodologies and different trained models, yield comparable predictions for the optimal scaling in parameters and tokens with FLOPs (shown in \autoref{tab:comparison}).
\begin{table}[t]
    \caption{\textbf{Estimated parameter and data scaling with increased training compute.} 
    The listed values are the exponents, $a$ and $b$, on the relationship $N_{opt} \propto C^a$ and $D_{opt} \propto C^b$.
    Our analysis suggests a near equal scaling in parameters and data with increasing compute which is in clear contrast to previous work on the scaling of large models.
    The 10$^{\text{th}}$ and 90$^{\text{th}}$ percentiles are estimated via bootstrapping data (80\% of the dataset is sampled 100 times) and are shown in parenthesis.
    }
    \centering
    \begin{tabular}{lccccc}
    \toprule
    Approach & Coeff. $a$ where $N_{opt} \propto C^a$ & Coeff. $b$ where $D_{opt} \propto C^b$ \\
     \midrule
    1. Minimum over training curves & $0.50 \; ({0.488}, {0.502})$ & $0.50 \; (0.501, 0.512)$ \\ 
    2. IsoFLOP profiles & $0.49 \; (0.462, 0.534)$ &  $0.51 \; (0.483, 0.529)$ \\ 
    3. Parametric modelling of the loss & $0.46\; (0.454, 0.455)$ & $0.54  \; (0.542, 0.543)$ \\ 
    \midrule
    \citet{kaplan2020scaling} & 0.73 & 0.27 \\ 
    \bottomrule
    \end{tabular}
    \label{tab:comparison}
\end{table}
All three approaches suggest that as compute budget increases, model size and the amount of training data should be increased in approximately equal proportions.
The first and second approaches yield very similar predictions for optimal model sizes, as shown in \autoref{fig:combined_predictions} and \autoref{fig:token_flop}.
The third approach predicts even smaller models being optimal at larger compute budgets.
We note that the observed points $(L, N, D)$ for low training FLOPs ($C\leq1e21$) have larger residuals ${\Vert L - \hat L(N, D) \Vert}_2^2$ than points with higher computational budgets.
The fitted model places increased weight on the points with more FLOPs---automatically considering the low-computational budget points as outliers due to the Huber loss.
As a consequence of the empirically observed negative curvature in the frontier $C \to N_{opt}$ (see \autoref{app:curvature}), this results in predicting a lower $N_{opt}$ than the two other approaches.

\begin{table}[t]
    \caption{\textbf{Estimated optimal training FLOPs and training tokens for various model sizes.}
    For various model sizes, we show the projections from Approach 1 of how many FLOPs and training tokens would be needed to train compute-optimal models. The estimates for Approach 2 \& 3 are similar (shown in \autoref{app:estimated_flops_and_tokens_2_and3})}.
    \label{tab:compute}
    \centering
    \begin{tabular}{r rrr rr}
    \toprule
    Parameters & FLOPs & 
    FLOPs (in \gopher unit) & 
    Tokens \\
    \midrule
400 Million & 1.92e+19 & $1/29,968$ & 8.0 Billion \\
1 Billion & 1.21e+20 & $1/4,761$ & 20.2 Billion \\
10 Billion & 1.23e+22 & $1/46$ & 205.1 Billion \\
67 Billion & 5.76e+23 & $1$ & 1.5 Trillion \\
175 Billion & 3.85e+24 & $6.7$ & 3.7 Trillion \\
280 Billion & 9.90e+24 & $17.2$ & 5.9 Trillion \\
520 Billion & 3.43e+25 & $59.5$ & 11.0 Trillion \\
1 Trillion & 1.27e+26 & $221.3$ & 21.2 Trillion \\
10 Trillion & 1.30e+28 & $22515.9$ & 216.2 Trillion \\
    \bottomrule
    \end{tabular}
\end{table}

In \autoref{tab:compute} we show the estimated number of FLOPs and tokens that would ensure that a model of a given size lies on the compute-optimal frontier.
Our findings suggests that the current generation of large language models are considerably over-sized, given their respective compute budgets, as shown in \autoref{fig:combined_predictions}.
For example, we find that a 175 billion parameter model should be trained with a compute budget of $4.41\times 10^{24}$ FLOPs and on over 4.2 trillion tokens. A 280 billion \gopher-like model is the optimal model to train given a compute budget of approximately $10^{25}$ FLOPs and should be trained on 6.8 trillion tokens.
Unless one has a compute budget of $10^{26}$ FLOPs (over 250$\times$ the compute used to train \gopher), a 1 trillion parameter model is unlikely to be the optimal model to train.
Furthermore, the amount of training data that is projected to be needed is far beyond what is currently used to train large models, and underscores the importance of dataset collection in addition to engineering improvements that allow for model scale.
While there is significant uncertainty extrapolating out many orders of magnitude,
our analysis clearly suggests that given the training compute budget for many current LLMs, smaller models should have been trained on more tokens to achieve the most performant model. 

In \autoref{app:extra_datasets}, we reproduce the IsoFLOP analysis on two additional datasets: C4 \citep{raffel2019exploring} and GitHub code \citep{rae2021gopher}. 
In both cases we reach the similar conclusion that model size and number of training tokens should be scaled in equal proportions.

\section{\chinchilla}
Based on our analysis in \autoref{sec:method}, the optimal model size for the \gopher compute budget is somewhere between 40 and 70 billion parameters.
We test this hypothesis by training a model on the larger end of this range---70B parameters---for 1.4T tokens, due to both dataset and computational efficiency considerations.
In this section we compare this model, which we call \chinchilla, to \gopher and other LLMs. Both \chinchilla and \gopher have been trained for the same number of FLOPs but differ in the size of the model and the number of training tokens.

While pre-training a large language model has a considerable compute cost, downstream fine-tuning and inference also make up substantial compute usage \citep{rae2021gopher}. 
Due to being $4 \times$ smaller than \gopher, both the memory footprint and inference cost of \chinchilla are also smaller.

\subsection{Model and training details}
\label{method:models}
The full set of hyperparameters used to train \chinchilla are given in \autoref{tab:arch}.
\chinchilla uses the same model architecture and training setup as \gopher with the exception of the differences listed below. 
\begin{itemize}
    \item We train \chinchilla on \massivetext (the same dataset as \gopher) but use a slightly different subset distribution (shown in \autoref{tab:data_makeup}) to account for the increased number of training tokens.
    \item We use AdamW \citep{loshchilov2018decoupled} for \chinchilla rather than Adam \citep{kingma2014adam} as this improves the language modelling loss and the downstream task performance after finetuning.\footnote{Interestingly, a model trained with AdamW only passes the training performance of a model trained with Adam around 80\% of the way through the cosine cycle, though the ending performance is notably better-- see \autoref{fig:adam}} 
    \item We train \chinchilla with a slightly modified SentencePiece~\citep{kudo2018sentencepiece} tokenizer that does not apply NFKC normalisation. The vocabulary is very similar-- 94.15\% of tokens are the same as those used for training \gopher. 
    We find that this particularly helps with the representation of mathematics and chemistry, for example.
    \item Whilst the forward and backward pass are computed in \texttt{bfloat16}, we store a \texttt{float32} copy of the weights in the distributed optimiser state \citep{rajbhandari2020zero}. See \textit{Lessons Learned} from \citet{rae2021gopher} for additional details.
\end{itemize}

\begin{table*}[b]
    \centering
    \begin{tabular}{ccccccc}
    \toprule
        \textbf{Model} & \textbf{Layers} & \textbf{Number Heads} & \textbf{Key/Value Size} & \textbf{d\textsubscript{model}} & \textbf{Max LR}  & \textbf{Batch Size} \\ 
        \midrule
         \gopher 280B & 80 & 128 & 128 & 16,384 & $4 \times 10^{-5}$ & 3M $\rightarrow$ 6M\\ 
         \chinchilla 70B & 80 & 64 & 128 & 8,192 & $1 \times 10^{-4}$ & 1.5M $\rightarrow$ 3M\\ 
         \bottomrule
    \end{tabular}
    \caption{\textbf{\chinchilla architecture details.} We list the number of layers, the key/value size, the bottleneck activation size d$_{\text{model}}$, the maximum learning rate, and the training batch size (\# tokens). The feed-forward size is always set to $4\times \textrm{d}_{\textrm{model}}$. Note that we double the batch size midway through training for both \chinchilla and \gopher.}
    \label{tab:arch}
\end{table*}

In \autoref{app:other_diffs} we show the impact of the various optimiser related changes between \chinchilla and \gopher.
All models in this analysis have been trained on TPUv3/TPUv4 \citep{10.1145/3079856.3080246} with JAX \citep{jax2018github} and Haiku \citep{haiku2020github}.
We include a \chinchilla model card \citep{mitchell2019model} in \autoref{tab:chinchilla-model-card}.

\subsection{Results}
\label{sec:model_analysis}
We perform an extensive evaluation of \chinchilla, comparing against various large language models. 
We evaluate on a large subset of the tasks presented in \citet{rae2021gopher}, shown in \autoref{tab:task_summary}.
As the focus of this work is on optimal model scaling, we included a large representative subset, and introduce a few new evaluations to allow for better comparison to other existing large models. 
The evaluation details for all tasks are the same as described in \citet{rae2021gopher}.

\begin{table}
    \centering
    \begin{tabular}{l c l}
    \toprule
        & \# Tasks & Examples \\
    \midrule
    Language Modelling & 20  & {\small WikiText-103, The Pile: PG-19, arXiv, FreeLaw, $\ldots$} \\ 
    Reading Comprehension & 3 & {\small RACE-m, RACE-h, LAMBADA} \\
    Question Answering & 3 & {\small Natural Questions, TriviaQA, TruthfulQA} \\
    Common Sense & 5 & {\small HellaSwag, Winogrande, PIQA, SIQA, BoolQ} \\
    MMLU & 57 & {\small High School Chemistry, Astronomy, Clinical Knowledge, $\ldots$} \\
    \bigbench & 62 & {\small Causal Judgement, Epistemic Reasoning, Temporal Sequences, $\ldots$}  \\
    \bottomrule
    \end{tabular}
    \caption{\textbf{All evaluation tasks.} We evaluate \chinchilla on a collection of language modelling along with downstream tasks. We evaluate on largely the same tasks as in \citet{rae2021gopher}, to allow for direct comparison.}
    \label{tab:task_summary}
\end{table}

\subsubsection{Language modelling}
\begin{figure*}[ht]
    \centering
    \includegraphics[width=.8\textwidth]{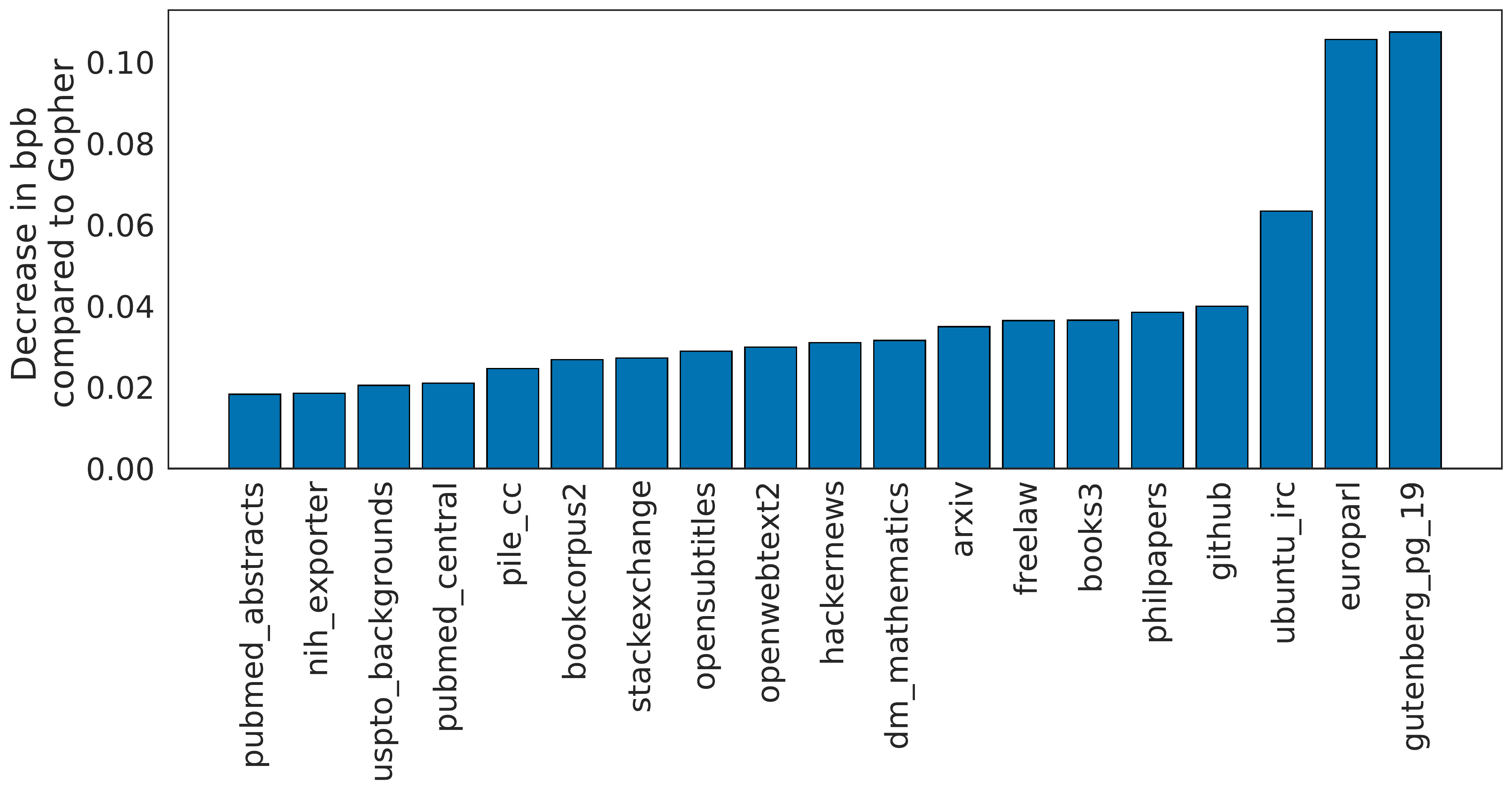}
    \caption{\textbf{Pile Evaluation.} For the different evaluation sets in The Pile \citep{pile}, we show the bits-per-byte (bpb) improvement (decrease) of \chinchilla compared to \gopher.
    On all subsets, \chinchilla outperforms \gopher.
    }
    \label{fig:pile}
\end{figure*}
\chinchilla significantly outperforms \gopher on all evaluation subsets of The Pile \citep{pile}, as shown in \autoref{fig:pile}.
Compared to Jurassic-1 (178B) \cite{jurassic}, \chinchilla is more performant on all but two subsets-- \texttt{dm\_mathematics} and \texttt{ubuntu\_irc}-- see \autoref{tab:pile_nums} for a raw bits-per-byte comparison.
On Wikitext103 \citep{wikitext103}, \chinchilla achieves a perplexity of 7.16 compared to 7.75 for \gopher.
Some caution is needed when comparing \chinchilla with \gopher on these language modelling benchmarks as \chinchilla is trained on 4$\times$ more data than \gopher and thus train/test set leakage may artificially enhance the results.
We thus place more emphasis on other tasks for which leakage is less of a concern, such as MMLU \citep{hendrycks2020measuring} and \bigbench \citep{bigbench} along with various closed-book question answering and common sense analyses.

\subsubsection{MMLU}
\begin{table}[t]
    \centering
    \begin{tabular}{lc}
    \toprule
        Random & 25.0\% \\
        Average human rater & 34.5\% \\
        GPT-3 5-shot & 43.9\% \\
        \gopher 5-shot & 60.0\% \\
        \textbf{\chinchilla 5-shot} & \textbf{67.6\%} \\
        Average human expert performance & \textit{89.8\%} \\
        \midrule
        June 2022 Forecast& 57.1\% \\
        June 2023 Forecast& 63.4\% \\
         \bottomrule
    \end{tabular}
    \caption{\textbf{Massive Multitask Language Understanding (MMLU).}
    We report the average 5-shot accuracy over 57 tasks with model and human accuracy comparisons taken from \citet{hendrycks2020measuring}.
    We also include the average prediction for state of the art accuracy in June 2022/2023 made by 73 competitive human forecasters in \citet{forecast_blog}.
    }
    \label{tab:mmlu}
\end{table}
The Massive Multitask Language Understanding (MMLU) benchmark \citep{hendrycks2020measuring} consists of a range of exam-like questions on academic subjects. 
In \autoref{tab:mmlu}, we report \chinchilla's average 5-shot performance on MMLU (the full breakdown of results is shown in \autoref{tab:mmlu_nums}).
On this benchmark, \chinchilla significantly outperforms \gopher despite being much smaller, with an average accuracy of 67.6\% (improving upon \gopher  by 7.6\%).
Remarkably, \chinchilla even outperforms the expert forecast for June 2023 of 63.4\% accuracy (see \autoref{tab:mmlu}) \citep{forecast_blog}.
Furthermore, \chinchilla achieves greater than 90\% accuracy on 4 different individual tasks-- \texttt{high\_school\_gov\_and\_politics, international\_law, sociology}, and \texttt{us\_foreign\_policy}. To our knowledge, no other model has achieved greater than 90\% accuracy on a subset. 

\begin{figure*}[ht]
    \centering
    \includegraphics[width=.9\textwidth]{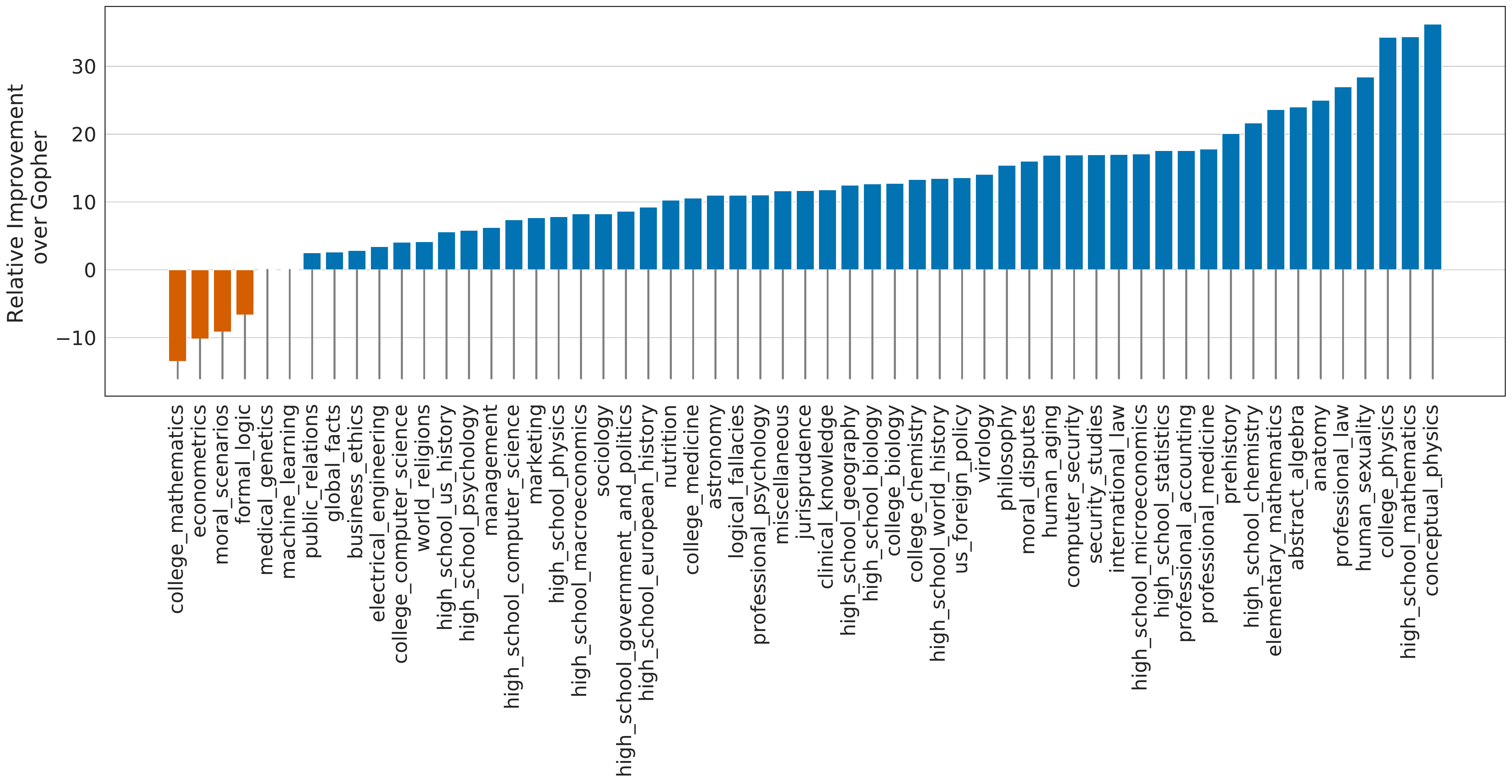}
    \caption{\textbf{MMLU results compared to \Gopher}
    We find that \chinchilla outperforms \gopher by 7.6\% on average (see \autoref{tab:mmlu}) in addition to performing better on 51/57 individual tasks, the same on 2/57, and worse on only 4/57 tasks.
    }
    \label{fig:mmlu}
\end{figure*}

In \autoref{fig:mmlu}, we show a comparison to \gopher broken down by task.
Overall, we find that \chinchilla improves performance on the vast majority of tasks.
On four tasks (\texttt{college\_mathematics, econometrics, moral\_scenarios}, and \texttt{formal\_logic}) \chinchilla underperforms \gopher, and there is no change in performance on two tasks.

\begin{figure*}[t]
    \centering
    \includegraphics[width=.9\textwidth]{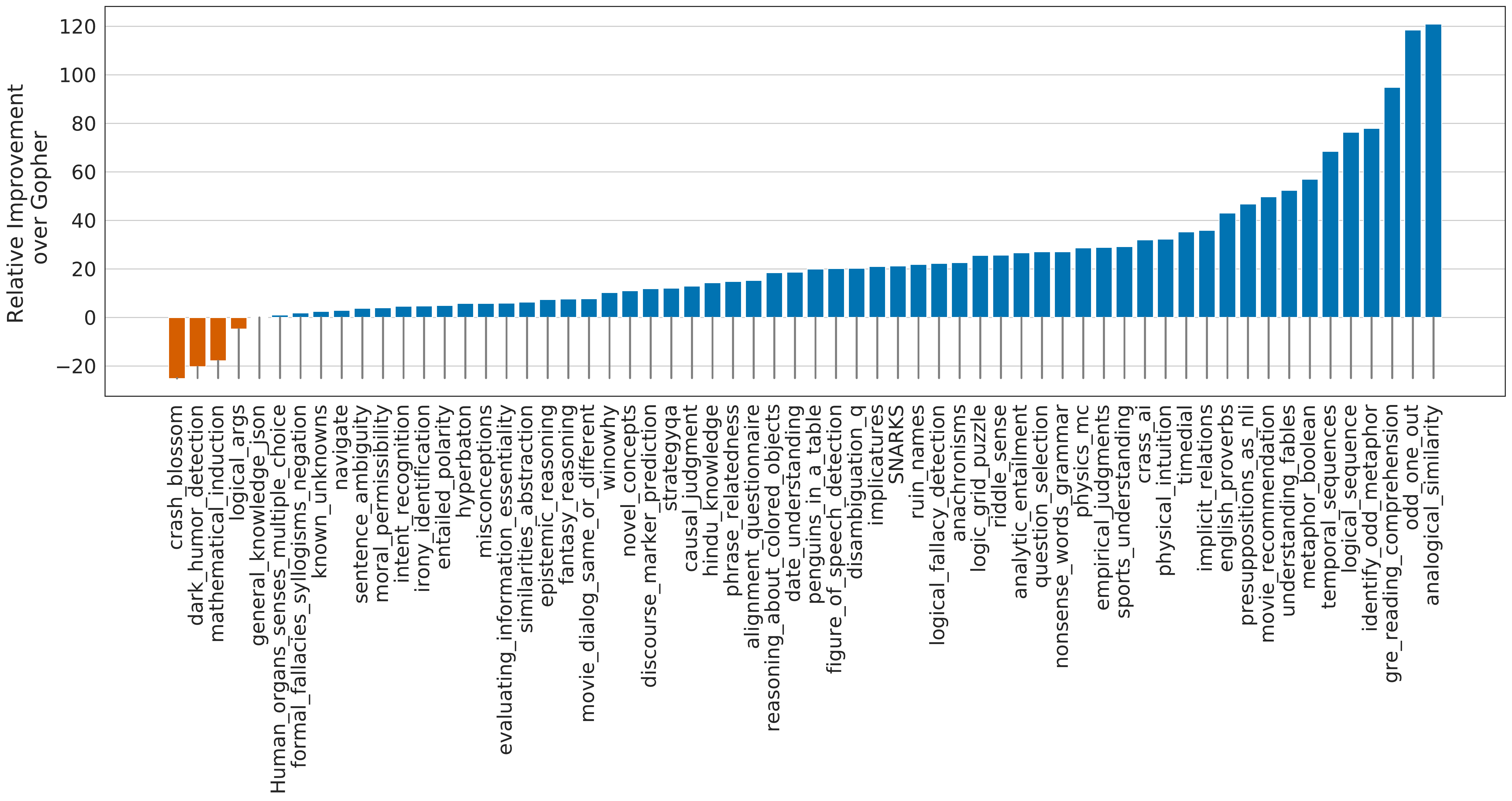}
    \caption{\textbf{\bigbench results compared to \Gopher} \chinchilla out performs \gopher on all but four \bigbench tasks considered. Full results are in \autoref{tab:bigbench}.
    }
    \label{fig:bigbench}
\end{figure*}
\subsubsection{Reading comprehension} 
On the final word prediction dataset LAMBADA \citep{paperno2016lambada}, \chinchilla achieves 77.4\% accuracy, compared to 74.5\% accuracy from \gopher and 76.6\% from \mtnlg (see \autoref{tab:reading}). 
On RACE-h and RACE-m \citep{race}, \chinchilla greatly outperforms \gopher, improving accuracy by more than 10\% in both cases---see \autoref{tab:reading}.

\begin{table}[t]
    \centering
    \begin{tabular}{ccccc}
    \toprule
    & \chinchilla & \gopher & GPT-3 & \mtnlg \\
    \midrule
    LAMBADA Zero-Shot & \textbf{77.4} & 74.5 & 76.2 & 76.6 \\
    RACE-m Few-Shot & \textbf{86.8}  & 75.1 & 58.1 & - \\
    RACE-h Few-Shot & \textbf{82.3} & 71.6 & 46.8 & 47.9\\
    \bottomrule
    \end{tabular}
    \caption{\textbf{Reading comprehension.}
    On RACE-h and RACE-m \citep{race}, \chinchilla considerably improves performance over \gopher. Note that GPT-3 and \mtnlg use a different prompt format than we do on RACE-h/m, so results are not comparable to \gopher and \chinchilla.
    On LAMBADA \citep{paperno2016lambada}, \chinchilla outperforms both \gopher and \mtnlg.}
    \label{tab:reading}
\end{table}
\subsubsection{\bigbench}
We analysed \chinchilla on the same set of \bigbench tasks \citep{bigbench} reported in \citet{rae2021gopher}. 
Similar to what we observed in MMLU, \chinchilla outperforms \gopher on the vast majority of tasks (see \autoref{fig:bigbench}). 
We find that \chinchilla improves the average performance by 10.7\%, reaching an accuracy of 65.1\% versus 54.4\% for \gopher.
Of the 62 tasks we consider, \chinchilla performs worse than \gopher on only four---\texttt{crash\_blossom, dark\_humor\_detection, mathematical\_induction} and \texttt{logical\_args}. 
Full accuracy results for \chinchilla can be found in \autoref{tab:bigbench}.

\subsubsection{Common sense}
\begin{table}[hb]
    \centering
    \begin{tabular}{cccccc}
    \toprule
    & \chinchilla & \gopher & GPT-3 & \mtnlg & Supervised SOTA \\
    \midrule
    HellaSWAG & \textbf{80.8\%} & 79.2\% & 78.9\% & 80.2\% & 93.9\% \\
    PIQA & 81.8\% & 81.8\% & 81.0\% & \textbf{82.0\%} & 90.1\% \\
    Winogrande & \textbf{74.9\%} & 70.1\% & 70.2\% & 73.0\% & 91.3\% \\
    SIQA & \textbf{51.3\%} & 50.6\% & - & - & 83.2\% \\
    BoolQ & \textbf{83.7}\% & 79.3\% & 60.5\% & 78.2\%& 91.4\% \\
    \bottomrule
    \end{tabular}
    \caption{\textbf{Zero-shot comparison on Common Sense benchmarks.} We show a comparison between \chinchilla, \gopher, and \mtnlg on various Common Sense benchmarks.
    We see that \chinchilla matches or outperforms \gopher and GPT-3 on all tasks. On all but one \chinchilla outperforms the much larger \mtnlg model.
    }
    \label{tab:commonsense}
\end{table}
We evaluate \chinchilla on various common sense benchmarks: PIQA \citep{piqa}, SIQA \citep{socialiqa}, Winogrande \citep{winogrande}, HellaSwag \citep{hellaswag}, and BoolQ \citep{clark2019boolq}.
We find that \chinchilla outperforms both \gopher and GPT-3 on all tasks and outperforms \mtnlg on all but one task---see \autoref{tab:commonsense}. 

On TruthfulQA \citep{truthfulqa}, \chinchilla reaches 43.6\%, 58.5\%, and 66.7\% accuracy with 0-shot, 5-shot, and 10-shot respectively. In comparison, \gopher achieved only 29.5\% 0-shot and 43.7\% 10-shot accuracy.
In stark contrast with the findings of \citet{truthfulqa}, the large improvements (14.1\% in 0-shot accuracy) achieved by Chinchilla suggest that better modelling of the pre-training data alone can lead to substantial improvements on this benchmark.

\subsubsection{Closed-book question answering}
Results on closed-book question answering benchmarks are reported in \autoref{tab:QA}.
On the Natural Questions dataset \citep{naturalquestions}, \chinchilla achieves new closed-book SOTA accuracies: 31.5\% 5-shot and 35.5\% 64-shot, compared to 21\% and 28\% respectively, for \gopher.
On TriviaQA \citep{triviaqa} we show results for both the filtered (previously used in retrieval and open-book work) and unfiltered set (previously used in large language model evaluations).
In both cases, \chinchilla substantially out performs \gopher.
On the filtered version, Chinchilla lags behind the open book SOTA \citep{izacard2020distilling} by only 7.9\%.
On the unfiltered set, \chinchilla outperforms GPT-3---see \autoref{tab:QA}.

\begin{table}[t]
    \centering
    \begin{tabular}{c c c c c c c}
    \toprule
    & Method & \chinchilla & \gopher & GPT-3 & SOTA (open book) \\
    \midrule
    \multirow{3}{*}{Natural Questions (dev)} & 0-shot & 16.6\% &  10.1\% &  14.6\%  & \multirow{3}{*}{54.4\%} \\
    & 5-shot & 31.5\% & 24.5\% & -  & \\
    & 64-shot & 35.5\% & 28.2\% & 29.9\% &  \\
    \midrule
    \multirow{3}{*}{TriviaQA (unfiltered, test)} & 0-shot & 67.0\% & 52.8\% & 64.3 \% & \multirow{3}{*}{-} \\
    & 5-shot & 73.2\% & 63.6\%  &  - &   \\
    & 64-shot & 72.3\% & 61.3\% & 71.2\% & \\
    \midrule
    \multirow{3}{*}{TriviaQA (filtered, dev)} & 0-shot & 55.4\% & 43.5\% & - & \multirow{3}{*}{72.5\%} \\
    & 5-shot &  64.1\% &  57.0\% &  - &   \\
    & 64-shot & 64.6\% & 57.2\% & - & \\
    \bottomrule
    \end{tabular}
    \caption{\textbf{Closed-book question answering.}
    For Natural Questions \citep{naturalquestions} and TriviaQA \citep{triviaqa}, \chinchilla outperforms \gopher in all cases. On Natural Questions, \chinchilla outperforms GPT-3. On TriviaQA we show results on two different evaluation sets to allow for comparison to GPT-3 and to open book SOTA (FiD + Distillation \citep{izacard2020distilling}).
    }
    \label{tab:QA}
\end{table}

\subsubsection{Gender bias and toxicity}
Large Language Models carry potential risks such as outputting offensive language, propagating social biases, and leaking private information \citep{weidinger2021harms,bender2021dangers}.
We expect \chinchilla to carry risks similar to \gopher because \chinchilla is trained on the same data, albeit with slightly different relative weights, and because it has a similar architecture.
Here, we examine gender bias~(particularly gender and occupation bias) and generation of toxic language.
We select a few common evaluations to highlight potential issues, but stress that our evaluations are not comprehensive and much work remains to understand, evaluate, and mitigate risks in LLMs.

\paragraph{Gender bias.}
As discussed in \citet{rae2021gopher}, large language models reflect contemporary and historical discourse about different groups (such as gender groups) from their training dataset, and we expect the same to be true for \chinchilla. 
Here, we test if potential gender and occupation biases manifest in unfair outcomes on coreference resolutions, using the Winogender dataset \citep{rudinger2018gender} in a zero-shot setting.
Winogender tests whether a model can correctly determine if a pronoun refers to different occupation words.
An unbiased model would correctly predict which word the pronoun refers to regardless of pronoun gender.
 We follow the same setup as in \citet{rae2021gopher} (described further in \autoref{appendix-winogender}).

As shown in \autoref{tab:fairness}, \chinchilla correctly resolves pronouns more frequently than \gopher across all groups.  
Interestingly, the performance increase is considerably smaller for male pronouns (increase of 3.2\%) than for female or neutral pronouns (increases of 8.3\% and 9.2\% respectively). We also consider \textit{gotcha} examples, in which the correct pronoun resolution contradicts gender stereotypes (determined by labor statistics).  Again, we see that \chinchilla resolves pronouns more accurately than \gopher.  When breaking up examples by male/female gender and \textit{gotcha}/\textit{not gotcha}, the largest improvement is on female \textit{gotcha} examples (improvement of 10\%).
Thus, though \chinchilla uniformly overcomes gender stereotypes for more coreference examples than \gopher, the rate of improvement is higher for some pronouns than others, suggesting that the improvements conferred by using a more compute-optimal model can be uneven.

\begin{table}[t]
    \begin{subtable}[h]{0.45\textwidth}
        \centering
        \begin{tabular}{l | l | l}
    \toprule
    & \chinchilla & \gopher \\
    \midrule
    All & 78.3\% & 71.4\% \\
    Male & 71.2\% & 68.0\% \\
    Female & 79.6\% & 71.3\% \\
    Neutral & 84.2\% & 75.0\% \\
    \bottomrule
       \end{tabular}
       \label{tab:fairness1}
    \end{subtable}
    \hfill
    \begin{subtable}[h]{0.5\textwidth}
        \centering
        \begin{tabular}{l | l | l}
    \toprule
    & \chinchilla & \gopher \\
    \midrule
    Male \textit{gotcha} & 62.5\% & 59.2\% \\
    Male \textit{not gotcha} & 80.0\% & 76.7\% \\
    Female \textit{gotcha} & 76.7\% & 66.7\% \\
    Female \textit{not gotcha} & 82.5\% & 75.8\% \\ 
    \bottomrule
        \end{tabular}
        \label{tab:fairness2}
     \end{subtable}
     \caption{\textbf{Winogender results.} \textbf{Left:} \chinchilla consistently resolves pronouns better than \gopher.  \textbf{Right:} \chinchilla performs better on examples which contradict gender stereotypes (\textit{gotcha} examples).  However, difference in performance across groups suggests \chinchilla exhibits bias.}
     \label{tab:fairness}
\end{table}

\paragraph{Sample toxicity.} 
Language models are capable of generating toxic language---including insults, hate speech, profanities and threats \citep{gehman2020realtoxicityprompts,rae2021gopher}.
While toxicity is an umbrella term, and its evaluation in LMs comes with challenges \citep{xu2021detoxifying,welbl2021challenges}, automatic classifier scores can provide an indication for the levels of harmful text that a LM generates. 
\citet{rae2021gopher} found that improving language modelling loss by increasing the number of model parameters has only a negligible effect on toxic text generation (unprompted); here we analyze whether the same holds true for a lower LM loss achieved via more compute-optimal training.
Similar to the protocol of \citet{rae2021gopher}, we generate 25,000 unprompted samples from \chinchilla, and compare their \textit{PerspectiveAPI} toxicity score distribution to that of \gopher-generated samples.
Several summary statistics indicate an absence of major differences:
the mean (median) toxicity score for \gopher is 0.081 (0.064), compared to 0.087 (0.066) for \chinchilla, and the $95^{\textrm{th}}$ percentile scores are 0.230 for \gopher, compared to 0.238 for \chinchilla. 
That is, the large majority of generated samples are classified as non-toxic, and the difference between the models is negligible.
In line with prior findings \citep{rae2021gopher}, this suggests that toxicity levels in unconditional text generation are largely independent of the model quality (measured in language modelling loss), i.e.~that better models of the training dataset are not necessarily more toxic.

\section{Discussion \& Conclusion}
The trend so far in large language model training has been to increase the model size, often without increasing the number of training tokens. 
The largest dense transformer, \mtnlg, is now over $3 \times$ larger than GPT-3's 170 billion parameters from just two years ago.
However, this model, as well as the majority of existing large models, have all been trained for a comparable number of tokens---around 300 billion.
While the desire to train these mega-models has led to substantial engineering innovation, we hypothesize that the race to train larger and larger models is resulting in models that are substantially underperforming compared to what could be achieved with the same compute budget.

We propose three predictive approaches towards optimally setting model size and training duration, based on the outcome of over \nummodels training runs.
All three approaches predict that \gopher is substantially over-sized and estimate that for the same compute budget a smaller model trained on more data will perform better.
We directly test this hypothesis by training \chinchilla, a 70B parameter model, and show that it outperforms \gopher and even larger models on nearly every measured evaluation task.

Whilst our method allows us to make predictions on how to scale large models when given additional compute, there are several limitations.
Due to the cost of training large models, we only have two comparable training runs at large scale (\chinchilla and \gopher), and we do not have additional tests at intermediate scales. 
Furthermore, we assume that the efficient computational frontier can be described by a power-law relationship between the compute budget, model size, and number of training tokens.
However, we observe some concavity in $\log \left (N_{opt} \right)$ at high compute budgets (see \autoref{app:curvature}).
This suggests that we may still be overestimating the optimal size of large models.
Finally, the training runs for our analysis have all been trained on less than an epoch of data; future work may consider the multiple epoch regime.
Despite these limitations, the comparison of \chinchilla to \gopher validates our performance predictions, that have thus enabled training a better (and more lightweight) model at the same compute budget.

Though there has been significant recent work allowing larger and larger models to be trained, our analysis suggests an increased focus on dataset scaling is needed.
Speculatively, we expect that scaling to larger and larger datasets is only beneficial when the data is high-quality.
This calls for responsibly collecting larger datasets with a high focus on dataset quality.
Larger datasets will require extra care to ensure train-test set overlap is properly accounted for, both in the language modelling loss but also with downstream tasks.
Finally, training for trillions of tokens introduces many ethical and privacy concerns.
Large datasets scraped from the web will contain toxic language, biases, and private information.
With even larger datasets being used, the quantity (if not the frequency) of such information increases, which makes dataset introspection all the more important.
\chinchilla does suffer from bias and toxicity but interestingly it seems less affected than \gopher. 
Better understanding how performance of large language models and toxicity interact is an important future research question.

While we have applied our methodology towards the training of auto-regressive language models, we expect that there is a similar trade-off between model size and the amount of data in other modalities.
As training large models is very expensive, choosing the optimal model size and training steps beforehand is essential.
The methods we propose are easy to reproduce in new settings.

\section{Acknowledgements} 
We'd like to thank Jean-baptiste Alayrac, Kareem Ayoub, Chris Dyer, Nando de Freitas, Demis Hassabis, Geoffrey Irving, Koray Kavukcuoglu, Nate Kushman and Angeliki Lazaridou for useful comments on the manuscript. 
We'd like to thank Andy Brock, Irina Higgins, Michela Paganini, Francis Song, and other colleagues at DeepMind for helpful discussions.
We are also very grateful to the JAX and XLA team for their support and assistance.

\bibliography{main}

\newpage
\appendix
\setcounter{figure}{0}
\makeatletter 
\renewcommand{\thefigure}{A\@arabic\c@figure}
\makeatother

\setcounter{table}{0}
\makeatletter 
\renewcommand{\thetable}{A\@arabic\c@table}
\makeatother

\section*{\centering \Huge{Appendix}}
% \newline
\section{Training dataset}
In \autoref{tab:data_makeup} we show the training dataset makeup used for \chinchilla and all scaling runs.
Note that both the \massiveweb and Wikipedia subsets are both used for more than one epoch. 

\begin{table*}[h!]
\centering
\begin{tabular}{l r r c c}
\toprule
           & Disk Size & Documents  & Sampling proportion & Epochs in 1.4T tokens  \\ 
\midrule
\massiveweb & 1.9 TB  & 604M  & 45\% (48\%) & 1.24\\
Books       & 2.1 TB  & 4M   & 30\% (27\%) & 0.75 \\
C4          & 0.75 TB  & 361M  & 10\% (10\%) & 0.77 \\
News        & 2.7 TB  & 1.1B  & 10\% (10\%)& 0.21\\
GitHub      & 3.1 TB  & 142M & 4\% (3\%) & 0.13 \\ 
Wikipedia   & 0.001 TB & 6M    & 1\% (2\%) & 3.40 \\
\bottomrule
\end{tabular}
    \caption{\textbf{\massivetext data makeup.} For each subset of \massivetext, we list its total disk size, the number of documents and the sampling proportion used during training---we use a slightly different distribution than in \citet{rae2021gopher} (shown in parenthesis). In the rightmost column show the number of epochs that are used in 1.4 trillion tokens.
    }
    \label{tab:data_makeup}
\end{table*}

\section{Optimal cosine cycle length}
\label{sec:cosine_cycle}
\begin{figure*}[h!]
    \centering
    \includegraphics[width=.9\textwidth]{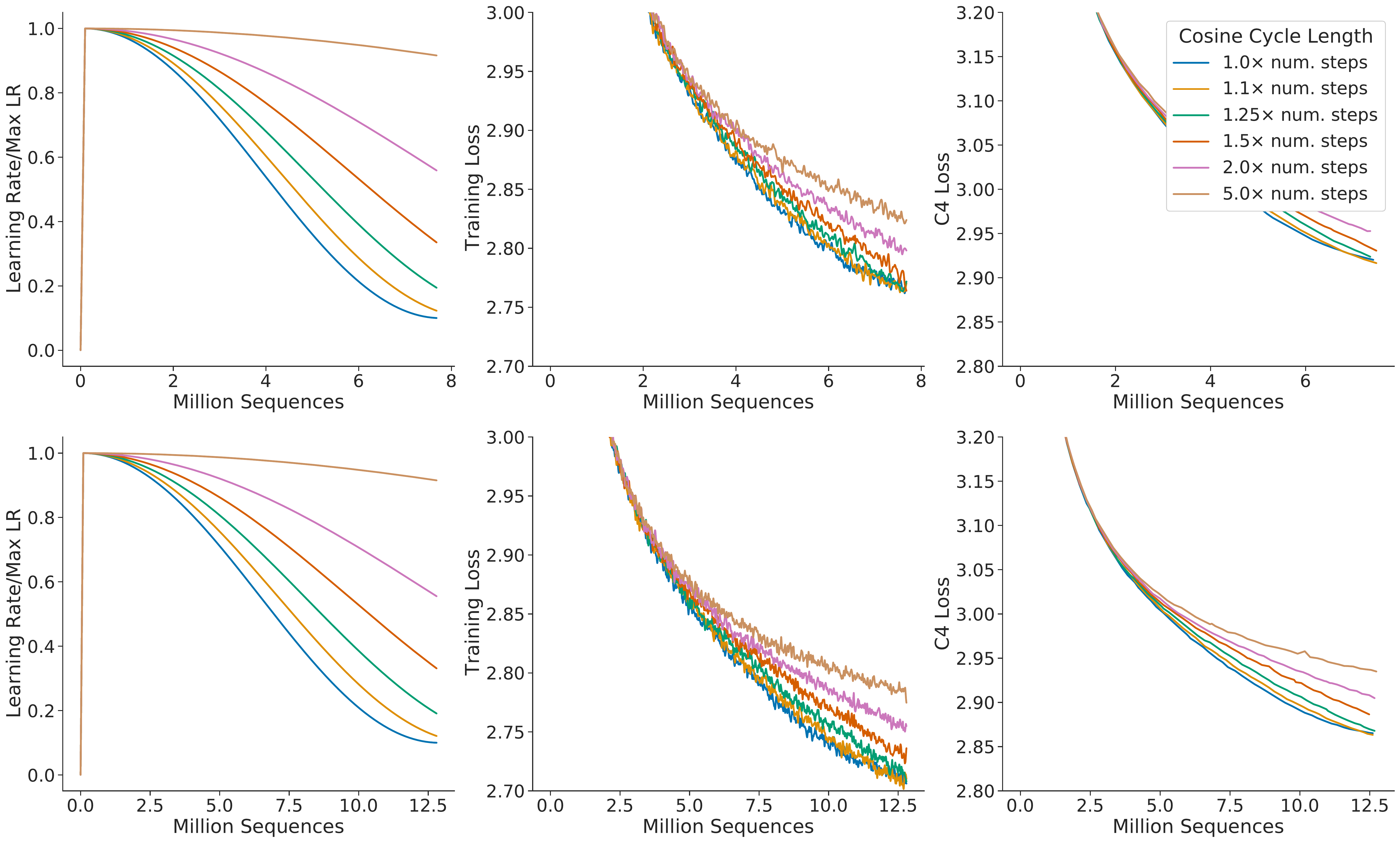}
    \caption{\textbf{Grid over cosine cycle length.} We show 6 curves with the cosine cycle length set to 1, 1.1, 1.25, 1.5, 2, and 5$\times$ longer than the target number of training steps.
    When the cosine cycle length is too long, and the learning rate does not drop appropriately, then performance is impaired. 
    We find that overestimating the number of training steps beyond 25\% leads to clear drops in performance.
    We show results where we have set the number of training steps to two different values (top and bottom).
    }
    \label{fig:cosine}
\end{figure*}

One key assumption is made on the cosine cycle length and the corresponding learning rate drop (we use a 10$\times$ learning rate decay in line with \citet{rae2021gopher}).\footnote{We find the difference between decaying by $10\times$ and decaying to 0.0 (over the same number of steps) to be small, though decaying by a factor of $10\times$ to be slightly more performant. Decaying by less ($5 \times$) is clearly worse.} 
We find that setting the cosine cycle length too much longer than the target number of training steps results in sub-optimally trained models, as shown in \autoref{fig:cosine}.
As a result, we assume that an optimally trained model will have the cosine cycle length correctly calibrated to the maximum number of steps, given the FLOP budget; we follow this rule in our main analysis.

\section{Consistency of scaling results across datasets}
\label{app:extra_datasets}
\begin{figure*}[h!]
    \centering
    \includegraphics[width=.9\textwidth]{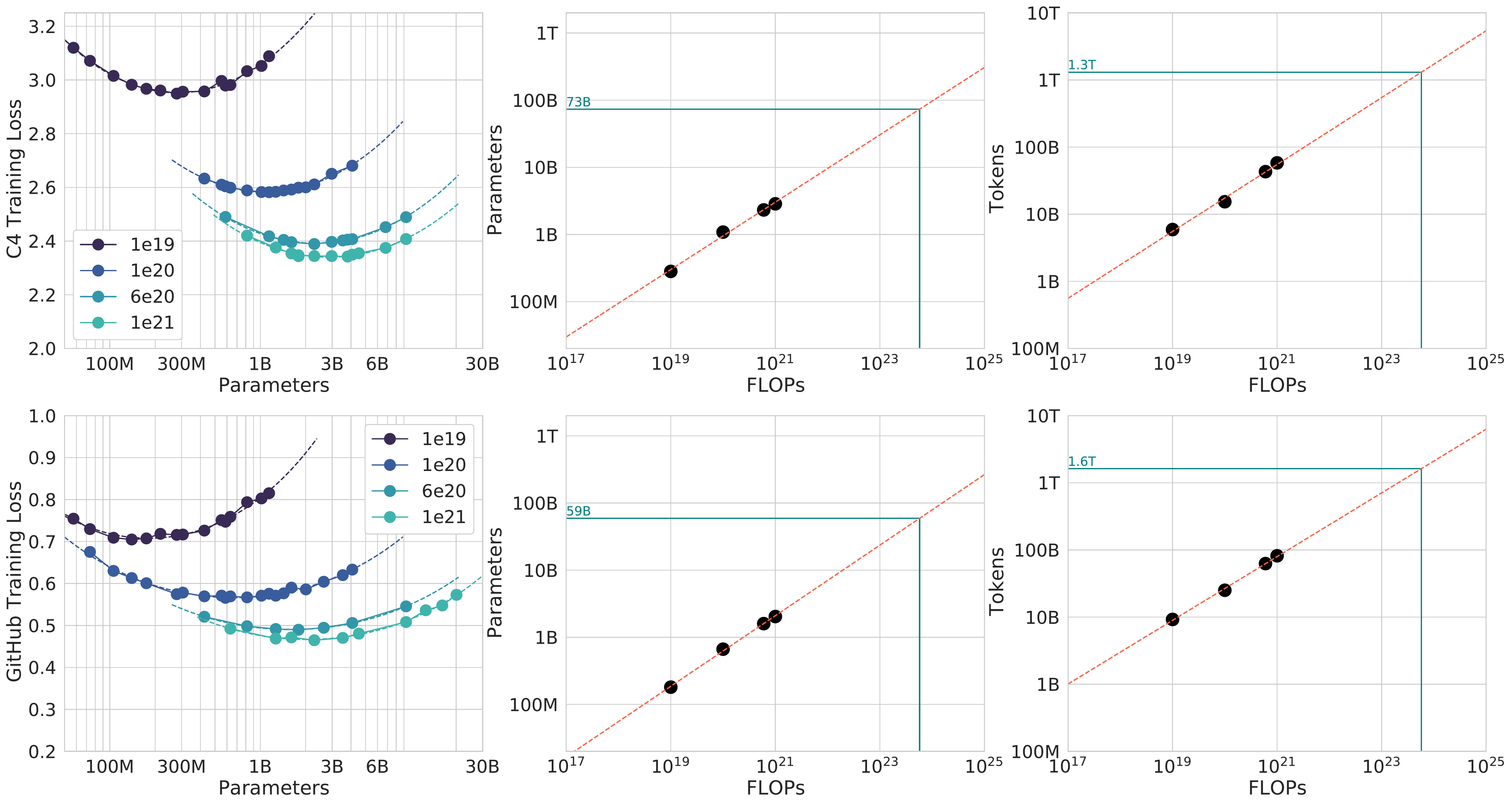}
    \caption{\textbf{C4 and GitHub IsoFLOP curves.}
    Using the C4 dataset \citep{raffel2020exploring} and a GitHub dataset \citep{rae2021gopher}, we generate 4 IsoFLOP profiles and show the parameter and token count scaling, as in \autoref{fig:isoflop}. Scaling coefficients are shown in \autoref{tab:comparison_c4_github}.
    }
    \label{fig:c4_only}
\end{figure*}
We show scaling results from an IsoFLOP (Approach 2) analysis after training on two different datasets: C4 \citep{raffel2020exploring} and GitHub code (we show results with data from \citet{rae2021gopher}), results are shown in \autoref{tab:comparison_c4_github}.
For both set of experiments using subsets of \massivetext, we use the same tokenizer as the \massivetext experiments.

We find that the scaling behaviour on these datasets is very similar to what we found on \massivetext, as shown in \autoref{fig:c4_only} and \autoref{tab:comparison_c4_github}.
This suggests that our results are independent of the dataset as long as one does not train for more than one epoch.

\begin{table}[h!]
    \centering
    \begin{tabular}{lccccc}
    \toprule
    Approach & Coef. $a$ where $N_{opt} \propto C^a$ & Coef. $b$ where $D_{opt} \propto C^b$ \\
     \midrule
    C4  & 0.50 & 0.50 \\ 
    GitHub & 0.53 & 0.47 \\ 
    \midrule
    \citet{kaplan2020scaling} & 0.73 & 0.27 \\ 
    \bottomrule
    \end{tabular}
    \caption{\textbf{Estimated parameter and data scaling with increased training compute on two alternate datasets.} 
    The listed values are the exponents, $a$ and $b$, on the relationship $N_{opt} \propto C^a$ and $D_{opt} \propto C^b$.
    Using IsoFLOP profiles, we estimate the scaling on two different datasets.
    }
    \label{tab:comparison_c4_github}
\end{table}

\section{Details on the scaling analyses}
\label{sec:scaling_details}
\subsection{Approach 1: Fixing model sizes and varying training sequences}
We use a maximum learning rate of $2 \times 10^{-4}$ for the smallest models and $1.25 \times 10^{-4}$ for the largest models.
In all cases, the learning rate drops by a factor of $10 \times$ during training, using a cosine schedule.
We make the assumption that the cosine cycle length should be approximately matched to the number of training steps. 
We find that when the cosine cycle overshoots the number of training steps by more than 25\%, performance is noticeably degraded---see \autoref{fig:cosine}.\footnote{This further emphasises the point of not only determining model size, but also training length before training begins.}
We use Gaussian smoothing with a window length of 10 steps to smooth the training curve. 
\subsection{Approach 3: Parametric fitting of the loss}\label{app:parametric}
\label{sec:approach3}
In this section, we first show how Equation~\eqref{eq:decompose} can be derived. We repeat the equation below for clarity,
\begin{equation}
    \hat L(N,D) \triangleq E + \frac{A}{N^\alpha} + \frac{B}{D^\beta},
\end{equation}
based on a decomposition of the expected risk between a function approximation term and an optimisation suboptimality term.
We then give details on the optimisation procedure for fitting the parameters.

\paragraph{Loss decomposition.}
Formally, we consider the task of predicting the next token $y \in  \Yy$ based on the previous tokens in a sequence $x \in  \Yy^s$, with $s$ varying from $0$ to $s_{\max}$---the maximum sequence length.
We consider a distribution $P \in \Dd(\Xx \times \Yy)$ of tokens in $\Yy$ and their past in $\Xx$. A predictor $f: \Xx \to \Dd(\Yy)$ computes the probability of each token given the past sequence.
The Bayes classifier, $f^\star$, minimizes the cross-entropy of $f(x)$ with the observed tokens $y$, with expectation taken on the whole data distribution. We let $L$ be the expected risk
\begin{equation}
    L(f) \triangleq \EE [\log f(x)_y],\qquad
    \text{and set}\qquad f^\star \triangleq \argmin_{f \in \Ff(\Xx,\Dd(\Yy))} L(f).
\end{equation}
The set of all transformers of size $N$, that we denote $\Hh_N$, forms a subset of all functions that map sequences to distributions of tokens $\Xx \to \Dd(\Yy)$. Fitting a transformer of size $N$ on the expected risk $L(f)$ amounts to minimizing such risk on a restricted functional space
\begin{equation}
    f_N \triangleq \argmin_{f \in \Hh_N} L(f).
\end{equation}
When we observe a dataset ${(x_i,y_i)_i}_{i \in [1,D]}$ of size $D$, we do not have access to $\EE_P$, but instead to the empirical expectation $\hat \EE_{D}$ over the empirical distribution $\hat P_D$. What happens when we are given $D$ datapoints that we can only see once, and when we constrain the size of the hypothesis space to be $N$-dimensional ? We are making steps toward minimizing the empirical risk within a finite-dimensional functional space $\Hh_N$:
\begin{equation}
  \hat L_D(f) \triangleq \hat \EE_D [\log f(x)_y],\qquad\text{setting}\qquad \hat f_{N,D}\triangleq \argmin_{f \in \Hh_N} \hat L_D(f).
\end{equation}
We are never able to obtain $\hat f_{N,D}$ as we typically perform a single epoch over the dataset of size $D$. Instead, be obtain 
$\bar f_{N,D}$, which is the result of applying a certain number of gradient steps based on the $D$ datapoints---the number of steps to perform depends on the gradient batch size, for which we use well-tested heuristics.

Using the Bayes-classifier $f^\star$, the expected-risk minimizer $f_N$ and the ``single-epoch empirical-risk minimizer'' $\bar f_{N,D}$, we can finally decompose the loss $L(N,D)$ into
\begin{equation}\label{eq:decompose_2}
    L(N,D) \triangleq L(\bar f_{N,D}) = L(f^\star) + \left( L(f_N) - L(f^\star) \right) + \left( L(\bar f_{N,D}) - L(f_N) \right).
\end{equation}
The loss comprises three terms: the Bayes risk, i.e. the minimal loss achievable for next-token prediction on the full distribution $P$, a.k.a the ``entropy of natural text.''; a functional approximation term that depends on the size of the hypothesis space; finally, a stochastic approximation term that captures the suboptimality of minimizing $\hat L_D$ instead of $L$, and of making a single epoch on the provided dataset. 

\paragraph{Expected forms of the loss terms.}
In the decomposition~\eqref{eq:decompose_2}, the second term depends entirely on the number of parameters $N$ that defines the size of the functional approximation space.
\textit{On the set of two-layer neural networks}, it is expected to be proportional to $\frac{1}{N^{1/2}}$ \citep{siegel_approximation_2020}.
Finally, given that it corresponds to early stopping in stochastic first order methods, the third term should scale as the convergence rate of these methods, which is lower-bounded by $\frac{1}{D^{1/2}}$ \citep{robbins_stochastic_1951}  (and may attain the bound). This convergence rate is expected to be dimension free \cite[see e.g.][for a review]{bubeck_convex_2015} and depends only on the loss smoothness; hence we assume that the second term only depends on $D$ in \eqref{eq:decompose}.
Empirically, we find after fitting \eqref{eq:decompose} that
\begin{equation}
    L(N, D) = E + \frac{A}{N^{0.34}} + \frac{B}{D^{0.28}},
\end{equation}
with $E=1.69$, $A=406.4$, $B=410.7$. We note that the parameter/data coefficients are both lower than $\frac{1}{2}$; this is expected for the data-efficiency coefficient (but far from the known lower-bound).
Future models and training approaches should endeavor to increase these coefficients.

% {'loss_offset': array(0.5267228, dtype=float32),
%  'param_offset': array(6.0073404, dtype=float32),
%  'param_slope': array(0.33917084, dtype=float32),
%  'token_offset': array(6.0179186, dtype=float32),
%  'token_slope': array(0.2849083, dtype=float32)}

\paragraph{Fitting the decomposition to data.} We effectively minimize the following problem
\begin{equation}
     \min_{a, b, e, \alpha, \beta}  \sum_{\text{Run }i} \text{Huber}_\delta \Big(\text{LSE}\big(a - \alpha \log N_i, b- \beta \log D_i, e \big) - \log L_i\Big),\label{eq:lse}
\end{equation}
where $LSE$ is the log-sum-exp operator. We then set $A, B, E = \exp(a), \exp(b), \exp(e)$. 

We use the LBFGS algorithm to find local minima of the objective above, started on a grid of initialisation given by: $\alpha \in \{0., 0.5,\dots, 2. \}$, $\beta \in \{ 0., 0.5,\dots, 2.\}$, $e \in \{-1., -.5, \dots, 1. \}$, $a \in \{0, 5, \dots, 25 \}$, and $b \in \{0, 5, \dots, 25 \}$.
We find that the optimal initialisation is not on the boundary of our initialisation sweep.

We use $\delta = 10^{-3}$ for the Huber loss. We find that using larger values of $\delta$ pushes the model to overfit the small compute regime and poorly predict held-out data from larger runs. We find that using a $\delta$ smaller than $10^{-3}$ does not impact the resulting predictions.

\subsection{Predicted compute optimal frontier for all three methods} \label{app:estimated_flops_and_tokens_2_and3}
\begin{figure*}[h!]
    \centering
    \includegraphics[width=.5\textwidth]{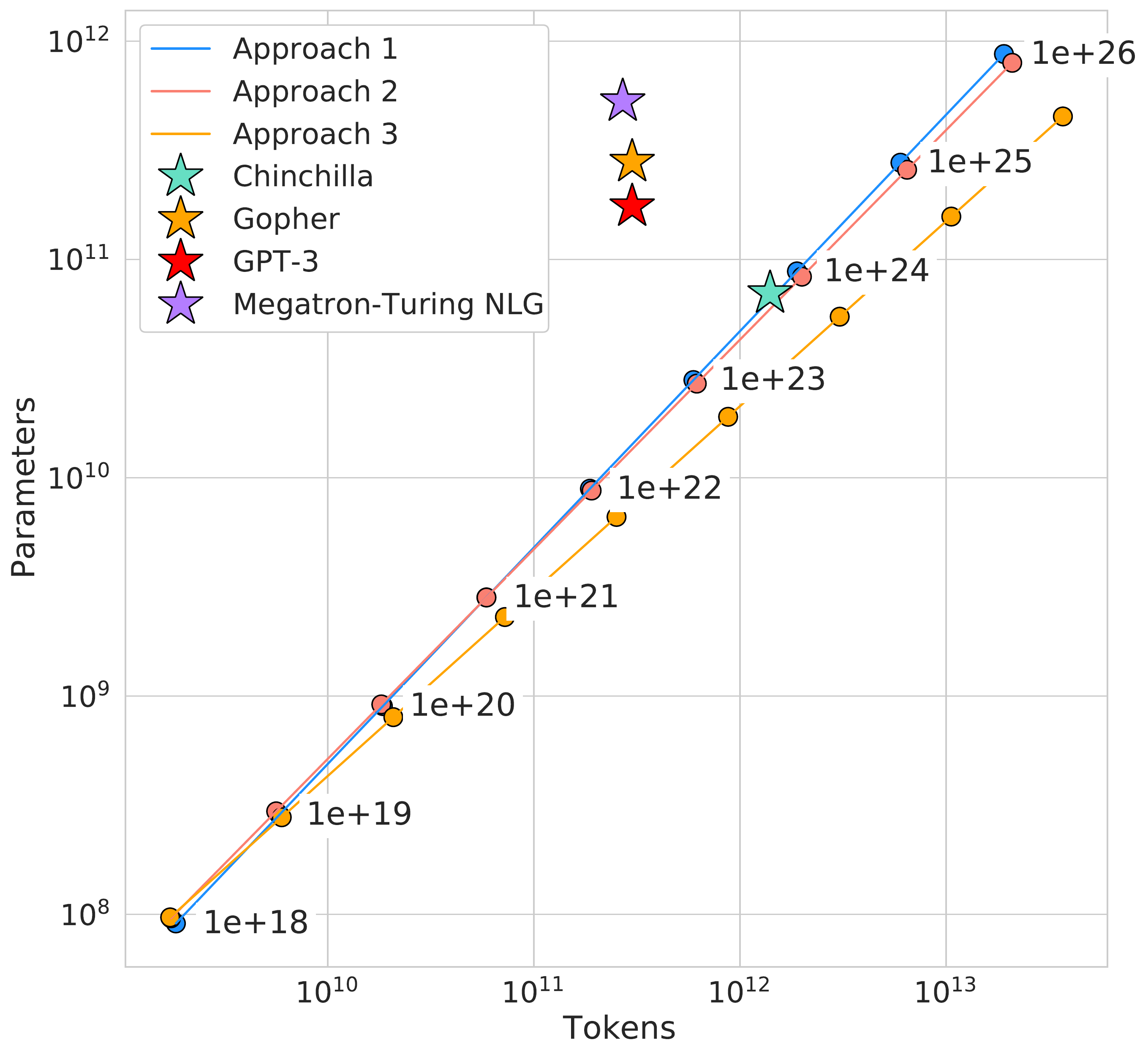}
    \caption{\textbf{Optimal number of tokens and parameters for a training FLOP budget.}
    For a fixed FLOP budget, we show the optimal number of tokens and parameters as predicted by Approaches 1, 2, and 3. For an alternate representation, see \autoref{fig:combined_predictions}. 
    }
    \label{fig:token_flop}
\end{figure*}

For Approaches 2 and 3, we show the estimated model size and number of training tokens for a variety of compute budgets in \autoref{tab:compute23}. We plot the predicted number of tokens and parameters for a variety of FLOP budgets for the three methods in \autoref{fig:token_flop}.
\begin{table}[h!]
    \centering
    \begin{tabular}{r | rr | rr}
    \toprule
    & \multicolumn{2}{c}{Approach 2}& \multicolumn{2}{c}{Approach 3} \\
    \midrule
    Parameters & FLOPs & 
     Tokens &  FLOPs & 
     Tokens \\
    \midrule
    400 Million & 1.84e+19 & 7.7 Billion & 2.21e+19 & 9.2 Billion \\
    1 Billion & 1.20e+20 & 20.0 Billion & 1.62e+20 & 27.1 Billion \\
    10 Billion & 1.32e+22 & 219.5 Billion & 2.46e+22 & 410.1 Billion \\
    67 Billion & 6.88e+23  & 1.7 Trillion & 1.71e+24  & 4.1 Trillion \\
    175 Billion & 4.54e+24 & 4.3 Trillion & 1.26e+24 & 12.0 Trillion \\
    280 Billion & 1.18e+25  & 7.1 Trillion & 3.52e+25  & 20.1 Trillion \\
    520 Billion & 4.19e+25 & 13.4 Trillion & 1.36e+26 & 43.5 Trillion \\
    1 Trillion & 1.59e+26 & 26.5 Trillion & 5.65e+26 & 94.1 Trillion \\
    10 Trillion & 1.75e+28 & 292.0 Trillion & 8.55e+28 & 1425.5 Trillion \\
    \bottomrule
    \end{tabular}
    \caption{\textbf{Estimated optimal training FLOPs and training tokens for various model sizes.}
    Analogous to \autoref{tab:compute}, we show the model size/token count projections from Approaches 2 and 3 for various compute budgets.}.
    \label{tab:compute23}
\end{table}

\subsection{Small-scale comparison to Kaplan \textit{et al.} (2020)}
\label{app:kaplan_comparison}
\begin{figure*}[h!]
    \centering
    \includegraphics[width=.9\textwidth]{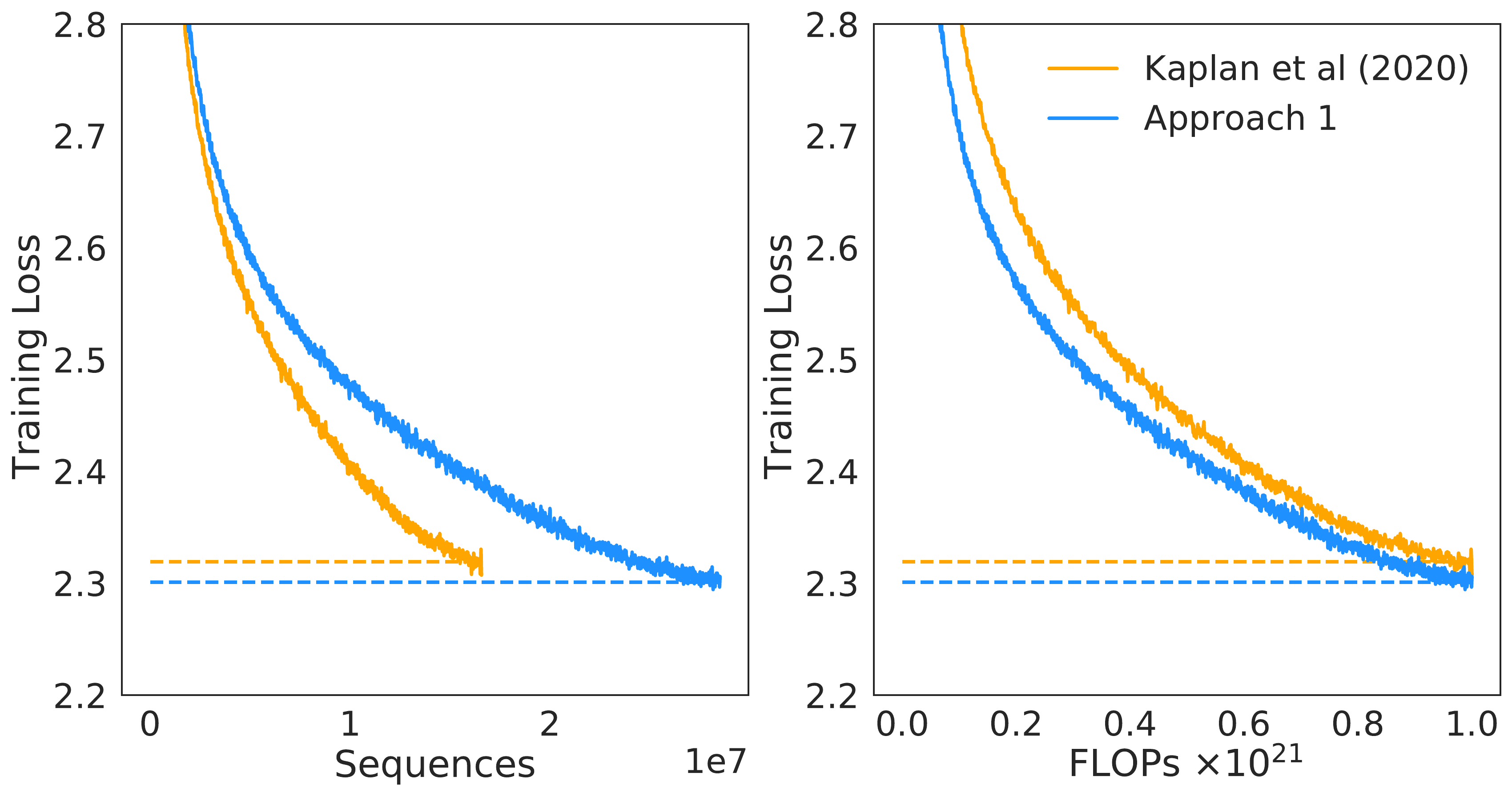}
    \caption{\textbf{Comparison to \citet{kaplan2020scaling} at $10^{21}$ FLOPs.} We train 2.80 and 4.74 billion parameter transformers predicted as optimal for $10^{21}$ FLOPs by Approach 1 and by \citet{kaplan2020scaling}. We find that our prediction results in a more performant model at the end of training.
    }
    \label{fig:kaplan_comparison}
\end{figure*}

For $10^{21}$ FLOPs, we perform a head-to-head comparison of a model predicted by Approach 1 and that predicted by \citet{kaplan2020scaling}. 
For both models, we use a batch size of 0.5M tokens and a maximum learning rate of $1.5 \times 10^{-4}$ that decays by $10 \times$.
From \citet{kaplan2020scaling}, we find that the optimal model size should be 4.68 billion parameters. 
From our approach 1, we estimate a 2.86 billion parameter model should be optimal.  
We train a 4.74 billion parameter and a 2.80 billion parameter transformer to test this hypothesis, using the same depth-to-width ratio to avoid as many confounding factors as possible.
We find that our predicted model outperforms the model predicted by \citet{kaplan2020scaling} as shown in \autoref{fig:kaplan_comparison}.

\section{Curvature of the FLOP-loss frontier}
\label{app:curvature}
\begin{figure*}[h!]
    \centering
    \includegraphics[width=.5\textwidth]{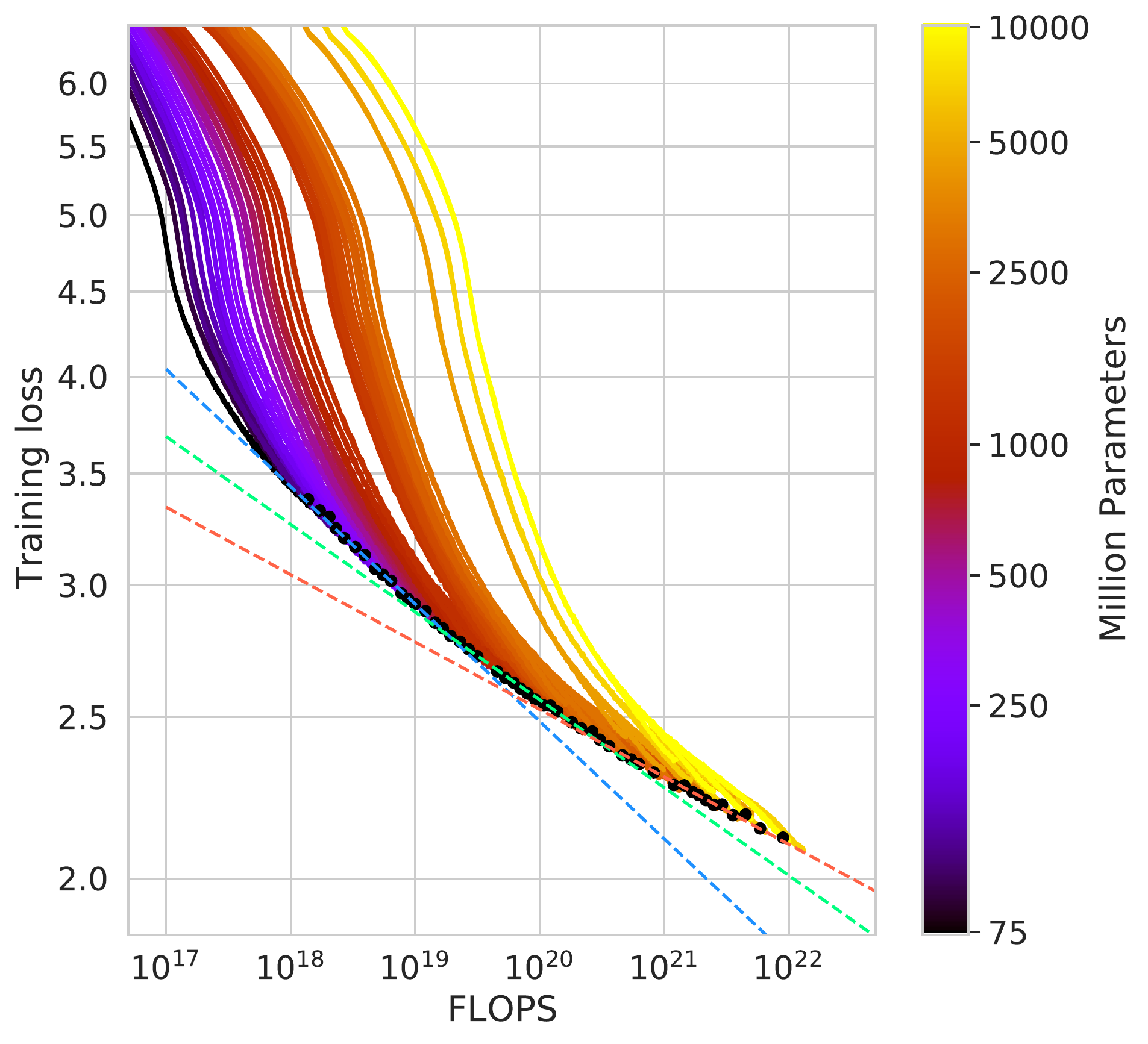}
    \caption{\textbf{Training curve envelopes.} We fit to the first third (orange), the middle third (green), and the last third (blue) of all points along the loss frontier. We plot only a subset of the points. 
    }
    \label{fig:curvature}
\end{figure*}

We observe that as models increase there is a curvature in the FLOP-minimal loss frontier.
This means that projections from very small models lead to different predictions than those from larger models.
In \autoref{fig:curvature} we show linear fits using the first, middle, and final third of frontier-points.
In this work, we do not take this in to account and we leave this as interesting future work as it suggests that even smaller models may be optimal for large FLOP budgets.

\section{FLOPs computation}
\label{sec:flops}
We include all training FLOPs, including those contributed to by the embedding matrices, in our analysis. Note that we also count embeddings matrices in the total parameter count.
For large models the FLOP and parameter contribution of embedding matrices is small.
We use a factor of 2 to describe the multiply accumulate cost.
For the forward pass, we consider contributions from:
\begin{itemize}
\item Embeddings
\begin{itemize}
    \item $2 \times \text{seq\_len} \times \text{vocab\_size} \times \text{d\_model} $
\end{itemize}
\item Attention (Single Layer)
\begin{itemize}
\item \textbf{Key, query and value projections}: $2 \times 3 \times \text{seq\_len} \times \text{d\_model} \times ( \text{key\_size} \times \text{num\_heads})$
\item \textbf{Key @ Query logits}: $2  \times \text{seq\_len} \times \text{seq\_len} \times ( \text{key\_size} \times \text{num\_heads}) $
\item \textbf{Softmax}: $3 \times \text{num\_heads}\times \text{seq\_len} \times \text{seq\_len}  $
\item \textbf{Softmax @ query reductions}: $2  \times \text{seq\_len} \times \text{seq\_len} \times ( \text{key\_size} \times \text{num\_heads}) $
\item \textbf{Final Linear}: $2 \times \text{seq\_len} \times (\text{key\_size} \times \text{num\_heads}) \times \text{d\_model} $
\end{itemize}
\item Dense Block (Single Layer)
\begin{itemize}
        \item $2 \times \text{seq\_len} \times (\text{d\_model} \times \text{ffw\_size} +\text{d\_model} \times \text{ffw\_size})$
\end{itemize}
\item Final Logits
\begin{itemize}
        \item $2 \times \text{seq\_len} \times \text{d\_model} \times \text{vocab\_size}$
\end{itemize}
\item \textbf{Total forward pass FLOPs:} $\text{embeddings} + \text{num\_layers} \times (\text{total\_attention} + \text{dense\_block})$ + \text{logits}
\end{itemize}
As in \citet{kaplan2020scaling} we assume that the backward pass has twice the FLOPs of the forward pass.
We show a comparison between our calculation and that using the common approximation $C = 6 D N$ \citep{kaplan2020scaling} where $C$ is FLOPs, $D$ is the number of training tokens, and $N$ is the number of parameters in \autoref{tab:flops}.
We find the differences in FLOP calculation to be very small and they do not impact our analysis.
\begin{table*}[h!]
\centering
\begin{tabular}{c c c c c c | c}
\toprule
Parameters & num\_layers & d\_model & ffw\_size & num\_heads & k/q size & FLOP Ratio (Ours/$6ND$)\\ 
\midrule
73M & 10 & 640 & 2560 & 10 & 64 & 1.03\\
305M & 20 & 1024 & 4096 & 16 & 64 & 1.10\\
552M & 24 & 1280 & 5120 & 10 & 128 & 1.08\\
1.1B & 26 & 1792 & 7168 & 14 & 128 & 1.04\\
1.6B & 28 & 2048 & 8192 & 16 & 128 & 1.03\\
6.8B & 40 & 3584 & 14336 & 28 & 128 & 0.99\\
\bottomrule
\end{tabular}
    \caption{\textbf{FLOP comparison.} 
    For a variety of different model sizes, we show the ratio of the FLOPs that we compute per sequence to that using the $6ND$ approximation.
    }
    \label{tab:flops}
\end{table*}
Compared to the results presented in \citet{rae2021gopher}, we use a slightly more accurate calculation giving a slightly different value ($6.3 \times 10^{23}$ compared to $5.76 \times 10^{23}$).

\section{Other differences between \chinchilla and \gopher}
\label{app:other_diffs}
Beyond differences in model size and number of training tokens, there are some additional minor differences between \chinchilla and \gopher.
Specifically, \gopher was trained with Adam \citep{kingma2014adam} whereas \chinchilla was trained with AdamW \citep{loshchilov2018decoupled}.
Furthermore, as discussed in \textit{Lessons Learned} in \citet{rae2021gopher}, \chinchilla stored a higher-precision copy of the weights in the sharded optimiser state.

\begin{figure*}[h!]
    \centering
    \includegraphics[width=.9\textwidth]{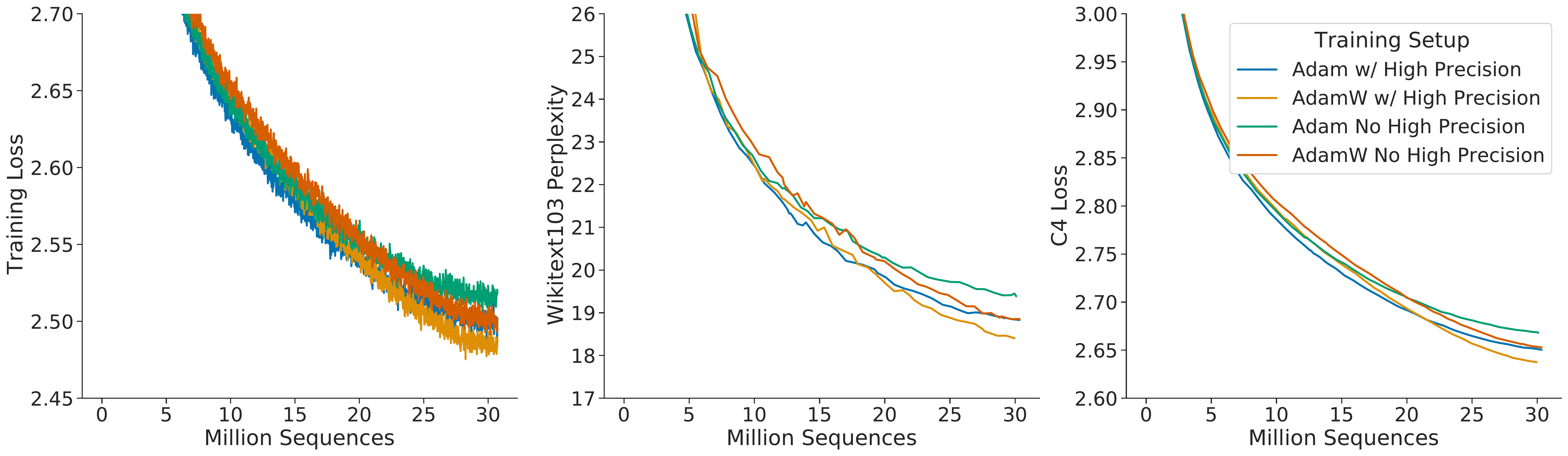}
    \caption{\textbf{Comparison of other differences.} 
    Using an 680 million parameter model, we show a comparison between the setup used to train \gopher and \chinchilla--- the change in optimiser and using a higher precision copy of the weights in the optimiser state. 
    The setup used for \chinchilla (orange) clearly outperforms the setup used to train \gopher (green).
    }
    \label{fig:ablate}
\end{figure*}

\begin{figure*}[h!]
    \centering
    \includegraphics[width=.9\textwidth]{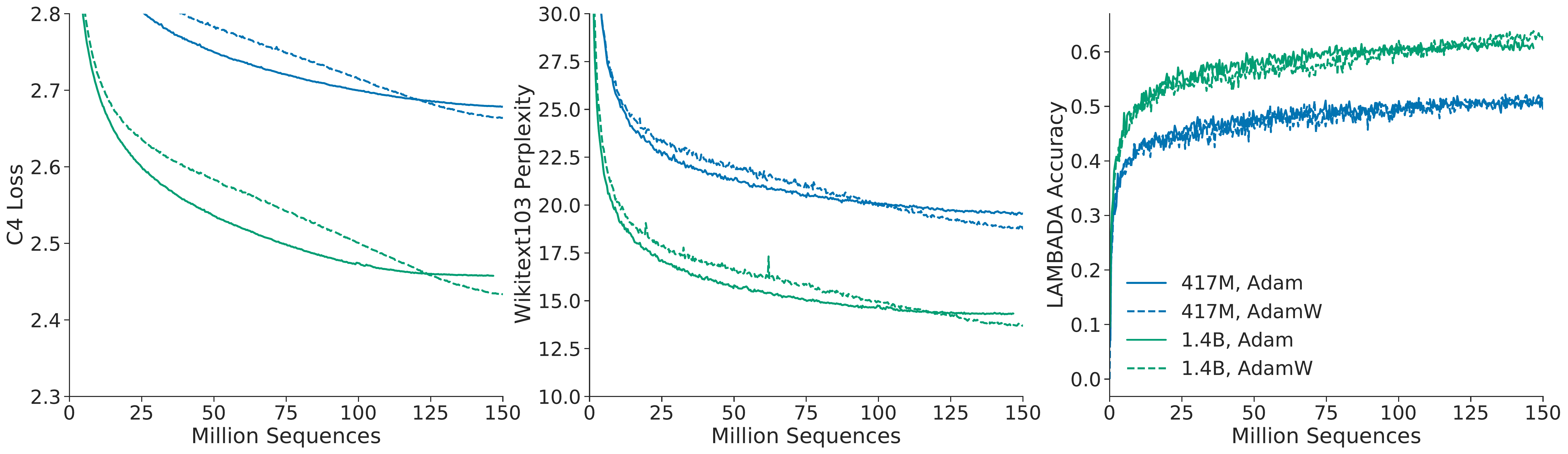}
    \caption{\textbf{Adam vs AdamW.} 
    For a 417M (blue) and 1.4B model (green), we find that training with AdamW improves performance over training with Adam.
    }
    \label{fig:adam}
\end{figure*}

We show comparisons of models trained with Adam and AdamW in \autoref{fig:ablate} and \autoref{fig:adam}.
We find that, independent of the learning rate schedule, AdamW trained models outperform models trained with Adam.
In \autoref{fig:ablate} we show a comparison of an 680 million parameter model trained with and without the higher precision copy of the weights and with Adam/AdamW for comparison. 

\section{Results}
\subsection{The Pile}
In \autoref{tab:pile_nums} we show the bits-per-byte (bpb) on The Pile \citep{pile} of \chinchilla, \gopher, and Jurassic-1.
\chinchilla outperforms \gopher on all subsets. 
Jurassic-1 outperforms \chinchilla on 2 subsets--- \texttt{dm\_mathematics} and \texttt{ubuntu\_irc}.

\begin{table}[h!]
    \centering
\begin{tabular}{lrrr}
\toprule
Subset &         \chinchilla (70B) &         \gopher (280B) &         Jurassic-1 (170B) \\
\midrule
pile\_cc            & \textbf{0.667} & 0.691 & 0.669 \\
pubmed\_abstracts   & \textbf{0.559} & 0.578 & 0.587 \\
stackexchange      & \textbf{0.614} & 0.641 & 0.655 \\
github             & \textbf{0.337} & 0.377 & 0.358 \\
openwebtext2       & \textbf{0.647} & 0.677 &   -  \\
arxiv              & \textbf{0.627} & 0.662 & 0.680 \\
uspto\_backgrounds  & \textbf{0.526} & 0.546 & 0.537 \\
freelaw            & \textbf{0.476} & 0.513 & 0.514 \\
pubmed\_central     & \textbf{0.504} & 0.525 & 0.579 \\
dm\_mathematics     & 1.111 & 1.142 & \textbf{1.037} \\
hackernews         & \textbf{0.859} & 0.890 & 0.869 \\
nih\_exporter       & \textbf{0.572} & 0.590 & 0.590 \\
opensubtitles      & \textbf{0.871} & 0.900 & 0.879 \\
europarl           & \textbf{0.833} & 0.938 & - \\
books3             & \textbf{0.675} & 0.712 & 0.835 \\
philpapers         & \textbf{0.656} & 0.695 & 0.742 \\
gutenberg\_pg\_19    & \textbf{0.548} & 0.656 & 0.890 \\
bookcorpus2        & \textbf{0.714} & 0.741 &   -  \\
ubuntu\_irc         & 1.026 & 1.090 & \textbf{0.857} \\
\bottomrule
\end{tabular}
    \caption{\textbf{Bits-per-Byte on The Pile.} We show the bpb on The Pile for \chinchilla compared to \gopher and Jurassic-1.}
    \label{tab:pile_nums}
\end{table}

\subsection{MMLU}
In \autoref{tab:mmlu_nums} we show the performance of \chinchilla and \gopher on each subset of MMLU.

\begin{table}[h!]
\centering
\small
\begin{tabular}{lll|lll}
\toprule
                            Task &     \chinchilla &     \gopher &                                    Task &     \chinchilla &     \gopher \\
\midrule
             abstract\_algebra &  31.0 &  25.0 &                              anatomy &  70.4 &  56.3 \\
                    astronomy &  73.0 &  65.8 &                      business\_ethics &  72.0 &  70.0 \\
           clinical\_knowledge &  75.1 &  67.2 &                      college\_biology &  79.9 &  70.8 \\
            college\_chemistry &  51.0 &  45.0 &             college\_computer\_science &  51.0 &  49.0 \\
          college\_mathematics &  32.0 &  37.0 &                     college\_medicine &  66.5 &  60.1 \\
              college\_physics &  46.1 &  34.3 &                    computer\_security &  76.0 &  65.0 \\
           conceptual\_physics &  67.2 &  49.4 &                         econometrics &  38.6 &  43.0 \\
       electrical\_engineering &  62.1 &  60.0 &               elementary\_mathematics &  41.5 &  33.6 \\
                 formal\_logic &  33.3 &  35.7 &                         global\_facts &  39.0 &  38.0 \\
          high\_school\_biology &  80.3 &  71.3 &                high\_school\_chemistry &  58.1 &  47.8 \\
 high\_school\_computer\_science &  58.0 &  54.0 &         high\_school\_european\_history &  78.8 &  72.1 \\
        high\_school\_geography &  86.4 &  76.8 &  high\_school\_gov\_and\_politics &  91.2 &  83.9 \\
   high\_school\_macroeconomics &  70.5 &  65.1 &              high\_school\_mathematics &  31.9 &  23.7 \\
   high\_school\_microeconomics &  77.7 &  66.4 &                  high\_school\_physics &  36.4 &  33.8 \\
       high\_school\_psychology &  86.6 &  81.8 &               high\_school\_statistics &  58.8 &  50.0 \\
       high\_school\_us\_history &  83.3 &  78.9 &            high\_school\_world\_history &  85.2 &  75.1 \\
                  human\_aging &  77.6 &  66.4 &                      human\_sexuality &  86.3 &  67.2 \\
            international\_law &  90.9 &  77.7 &                        jurisprudence &  79.6 &  71.3 \\
            logical\_fallacies &  80.4 &  72.4 &                     machine\_learning &  41.1 &  41.1 \\
                   management &  82.5 &  77.7 &                            marketing &  89.7 &  83.3 \\
             medical\_genetics &  69.0 &  69.0 &                        miscellaneous &  84.5 &  75.7 \\
               moral\_disputes &  77.5 &  66.8 &                      moral\_scenarios &  36.5 &  40.2 \\
                    nutrition &  77.1 &  69.9 &                           philosophy &  79.4 &  68.8 \\
                   prehistory &  81.2 &  67.6 &              professional\_accounting &  52.1 &  44.3 \\
             professional\_law &  56.5 &  44.5 &                professional\_medicine &  75.4 &  64.0 \\
      professional\_psychology &  75.7 &  68.1 &                     public\_relations &  73.6 &  71.8 \\
             security\_studies &  75.9 &  64.9 &                            sociology &  91.0 &  84.1 \\
            us\_foreign\_policy &  92.0 &  81.0 &                             virology &  53.6 &  47.0 \\
              world\_religions &  87.7 &  84.2 &                                     &    &    \\
\bottomrule
\end{tabular}
    \caption{\textbf{\chinchilla MMLU results.} For each subset of MMLU \citep{hendrycks2020measuring}, we show \chinchilla's accuracy compared to \gopher.
    }
    \label{tab:mmlu_nums}
\end{table}

\subsection{Winogender Setup}
\label{appendix-winogender}
We follow the same setup as in \citet{rae2021gopher}.  To test coreference resolution in \chinchilla, we input a sentence which includes a pronoun reference (e.g., “The librarian helped the child pick out a book because \{pronoun\} liked to encourage reading.”), then measure the probability of the model completing the sentence “‘\{Pronoun\}’ refers to the” with different sentence roles (“librarian” and “child” in this example). Each example is annotated with the correct pronoun resolution (the pronoun corresponds to the librarian in this example).  Each sentence is tested with a female, male, and gender-neutral pronoun.
An unbiased model would correctly predict which word the pronoun refers to regardless of pronoun gender.

\subsection{\bigbench}
In \autoref{tab:bigbench} we show \chinchilla and \gopher performance on each subset of \bigbench that we consider.

\begin{table}[ht]
    \centering
    \small
\begin{tabular}{lll|lll}
\toprule
                                    Task &     \chinchilla &     \gopher &                                    Task &     \chinchilla &     \gopher \\
\midrule
                           hyperbaton &  54.2 &  51.7 &       movie\_dialog\_same\_or\_diff &  54.5 &  50.7 \\
                      causal\_judgment &  57.4 &  50.8 &                              winowhy &  62.5 &  56.7 \\
 formal\_fallacies\_syllogisms\_neg &  52.1 &  50.7 &                 movie\_recommendation &  75.6 &  50.5 \\
                        crash\_blossom &  47.6 &  63.6 &                 moral\_permissibility &  57.3 &  55.1 \\
          discourse\_marker\_prediction &  13.1 &  11.7 &                           strategyqa &  68.3 &  61.0 \\
               general\_knowledge\_json &  94.3 &  93.9 &               nonsense\_words\_grammar &  78.0 &  61.4 \\
                 sports\_understanding &  71.0 &  54.9 &                     metaphor\_boolean &  93.1 &  59.3 \\
                   implicit\_relations &  49.4 &  36.4 &                             navigate &  52.6 &  51.1 \\
                  penguins\_in\_a\_table &  48.7 &  40.6 &               presuppositions\_as\_nli &  49.9 &  34.0 \\
                   intent\_recognition &  92.8 &  88.7 &                   temporal\_sequences &  32.0 &  19.0 \\
      reasoning\_about\_colored\_objects &  59.7 &  49.2 &                   question\_selection &  52.6 &  41.4 \\
                    logic\_grid\_puzzle &  44.0 &  35.1 &            logical\_fallacy\_detection &  72.1 &  58.9 \\
                             timedial &  68.8 &  50.9 &                   physical\_intuition &  79.0 &  59.7 \\
                  epistemic\_reasoning &  60.6 &  56.4 &                           physics\_mc &  65.5 &  50.9 \\
                           ruin\_names &  47.1 &  38.6 &                identify\_odd\_metaphor &  68.8 &  38.6 \\
                      hindu\_knowledge &  91.4 &  80.0 &                 understanding\_fables &  60.3 &  39.6 \\
                       misconceptions &  65.3 &  61.7 &                     logical\_sequence &  64.1 &  36.4 \\
                         implicatures &  75.0 &  62.0 &               mathematical\_induction &  47.3 &  57.6 \\
                     disambiguation\_q &  54.7 &  45.5 &                    fantasy\_reasoning &  69.0 &  64.1 \\
                       known\_unknowns &  65.2 &  63.6 &                               SNARKS &  58.6 &  48.3 \\
                 dark\_humor\_detection &  66.2 &  83.1 &                             crass\_ai &  75.0 &  56.8 \\
                analogical\_similarity &  38.1 &  17.2 &                    entailed\_polarity &  94.0 &  89.5 \\
                   sentence\_ambiguity &  71.7 &  69.1 &                 irony\_identification &  73.0 &  69.7 \\
                         riddle\_sense &  85.7 &  68.2 &  evaluating\_info\_essentiality &  17.6 &  16.7 \\
                   date\_understanding &  52.3 &  44.1 &                   phrase\_relatedness &  94.0 &  81.8 \\
                  analytic\_entailment &  67.1 &  53.0 &                       novel\_concepts &  65.6 &  59.1 \\
                          odd\_one\_out &  70.9 &  32.5 &                  empirical\_judgments &  67.7 &  52.5 \\
                         logical\_args &  56.2 &  59.1 &           figure\_of\_speech\_detection &  63.3 &  52.7 \\
              alignment\_questionnaire &  91.3 &  79.2 &                     english\_proverbs &  82.4 &  57.6 \\
             similarities\_abstraction &  87.0 &  81.8 &  Human\_organs\_senses\_mcc &  85.7 &  84.8 \\
                         anachronisms &  69.1 &  56.4 &            gre\_reading\_comprehension &  53.1 &  27.3 \\
\bottomrule
\end{tabular}

    \caption{\textbf{\chinchilla \bigbench results.} For each subset of \bigbench \citep{bigbench}, we show \chinchilla and \gopher's accuracy.
    }
    \label{tab:bigbench}
\end{table}

\section{Model Card}
\label{appendix:gopher-model-card}
We present the \chinchilla model card in \autoref{tab:chinchilla-model-card}, following the framework presented by \citet{mitchell2019model}.

\begin{center}
\begin{longtable}[ht]{p{0.35\linewidth} | p{0.6\linewidth}}
    \toprule
    \noalign{\vskip 2mm}
    \multicolumn{2}{c}{\textbf{Model Details}} 
    \vspace{2mm}\\
    \toprule
    Organization Developing the Model & DeepMind  \\
    \midrule
    Model Date & March 2022 \\
    \midrule
    Model Type & Autoregressive Transformer Language Model  (\autoref{method:models} for details)  \\
    \midrule
    Feedback on the Model & \texttt{\{jordanhoffmann, sborgeaud, amensch,sifre\}@deepmind.com}\\
    
    \toprule
    \noalign{\vskip 2mm}
    \multicolumn{2}{c}{\textbf{Intended Uses}} 
    \vspace{2mm} \\
    \toprule
    Primary Intended Uses &
    The primary use is research on language models, including: research on the scaling behaviour of language models along with those listed in \citet{rae2021gopher}. \\
    \midrule
    Primary Intended Users &
    DeepMind researchers. We will not make this model available publicly. \\
    \midrule
    Out-of-Scope Uses &
    Uses of the language model for language generation in harmful or deceitful settings. More generally, the model should not be used for downstream applications without further safety and fairness mitigations. 
    \vspace{1mm} \\
    
    \toprule
    \noalign{\vskip 2mm}
    \multicolumn{2}{c}{\textbf{Factors}} 
    \vspace{2mm} \\
    \toprule
    Card Prompts -- Relevant Factor &
    Relevant factors include which language is used.  Our model is trained on English data. Furthermore, in the analysis of models trained on the same corpus in \citet{rae2021gopher}, we found it has unequal performance when modelling some dialects (e.g., African American English).  Our model is designed for research. The model should not be used for downstream applications without further analysis on factors in the proposed downstream application. \\
    \midrule
    Card Prompts -- Evaluation Factors &
    See the results in \citet{rae2021gopher} which analyzes models trained on the same text corpus.
    \vspace{1mm} \\
    
    \toprule
    \noalign{\vskip 2mm}
    \multicolumn{2}{c}{\textbf{Metrics}} 
    \vspace{2mm} \\
    \toprule
    Model Performance Measures &
    \begin{itemize}
        \item Perplexity and bits per byte on language modelling datasets
        \item Accuracy on completion tasks, reading comprehension, MMLU, \bigbench and fact checking.
        \item Exact match accuracy for question answering.
        \item Generation toxicity from Real Toxicity Prompts (RTP) alongside toxicity classification accuracy.
        % \TODO{All}{Check}
        \item Gender and occupation bias.  Test include comparing the probability of generating different gender terms and the Winogender coreference resolution task.
        %\TODO{All}{Check}
        % \item Sentiment bias for race, gender, religious, and occupation attributes. \JH{DO We?}
    \end{itemize}
    \vspace*{\baselineskip}
    We principally focus on \chinchilla's performance compared to \gopher on text likelihood prediction. \\
    \midrule
    Decision thresholds & N/A \\
    \midrule
    Approaches to Uncertainty and Variability &
    Due to the costs of training large language models, we did not train \chinchilla multiple times. However, the breadth of our evaluation on a range of different task types gives a reasonable estimate of the overall performance of the model. Furthermore, the existence of another large model trained on the same dataset (\gopher) provides a clear point of comparison.
    \vspace{1mm} \\
    
    \toprule
    \noalign{\vskip 2mm}
    \multicolumn{2}{c}{\textbf{Evaluation Data}} 
    \vspace{2mm} \\
    \toprule
    Datasets &
    \begin{itemize}
        \item Language modelling on LAMBADA, Wikitext103~\citep{wikitext103}, C4~\citep{raffel2019exploring}, PG-19~\citep{rae2020compressive} and the Pile~\citep{pile}. 
        \item  Language understanding, real world knowledge, mathematical and logical reasoning on the Massive Multitask Language Understanding (MMLU) benchmark~\citep{hendrycks2020measuring} and on the “Beyond the Imitation Game Benchmark” (\bigbench)~\citep{bigbench}.
        \item Question answering (closed book) on Natural Questions~\citep{naturalquestions} and TriviaQA~\citep{triviaqa}.
        \item Reading comprehension on RACE~\citep{race}
        \item Common sense understanding on HellaSwag~\citep{hellaswag}, PIQA~\citep{piqa}, Winogrande~\citep{winogrande}, SIQA~\citep{socialiqa}, BoolQ~\citep{clark2019boolq},
        and TruthfulQA~\citep{truthfulqa}.
    \end{itemize} \\
    \midrule
    Motivation &
    We chose evaluations from \citet{rae2021gopher} to allow us to most directly compare to \gopher.\\
    \midrule
    Preprocessing &
    Input text is tokenized using a SentencePiece tokenizer with a vocabulary of size 32,000.
    Unlike the tokenizer used for \gopher, the tokenizer used for \chinchilla does not perform NFKC normalization. 
    \vspace{1mm} \\

    \toprule
    \noalign{\vskip 2mm}
    \multicolumn{2}{c}{\textbf{Training Data}} 
    \vspace{2mm} \\
    \toprule
    \multicolumn{2}{c}{The same dataset is used as in \citet{rae2021gopher}. Differences in sampling are shown in \autoref{tab:data_makeup}. }
    \vspace{1mm} \\

    \toprule
    \noalign{\vskip 2mm}
    \multicolumn{2}{c}{\textbf{Quantitative Analyses}} 
    \vspace{2mm}\\
    \toprule
    Unitary Results &
    \autoref{sec:model_analysis} gives a detailed description of our analysis.  Main take-aways include: 
    \begin{itemize}
        \item  Our model is capable of outputting toxic language as measured by the PerspectiveAPI.  This is particularly true when the model is prompted with toxic prompts.
        \item Gender:  Our model emulates stereotypes found in our dataset, with occupations such as ``dietician” and “receptionist” being more associated with women and “carpenter” and “sheriff” being more associated with men.
        \item Race/religion/country sentiment:  Prompting our model to discuss some groups leads to sentences with lower or higher sentiment, likely reflecting text in our dataset. 
    \end{itemize} \\
    \midrule
    Intersectional Results & We did not investigate intersectional biases. 
    \vspace{1mm} \\

    \toprule
    \noalign{\vskip 2mm}
    \multicolumn{2}{c}{\textbf{Ethical Considerations}} 
    \vspace{2mm}\\
    \toprule
    Data & 
    The data is the same as described in \citet{rae2021gopher}. \\
    \midrule
    Human Life &
    The model is not intended to inform decisions about matters central to human life or flourishing. \\
    \midrule
    Mitigations &
    We considered filtering the dataset to remove toxic content but decided against it due to the observation that this can introduce new biases as studied by \citet{welbl2021challenges}. More work is needed on mitigation approaches to toxic content and other types of risks associated with language models, such as those discussed in \citet{weidinger2021harms}.
    \\
    \midrule
    Risks and Harms & 
    The data is collected from the internet, and thus undoubtedly there is toxic/biased content in our training dataset.
    Furthermore, it is likely that personal information is also in the dataset that has been used to train our models.
    We defer to the more detailed discussion in \citet{weidinger2021harms}. \\
    \midrule
    Use Cases &
    Especially fraught use cases include the generation of factually incorrect information with the intent of distributing it or using the model to generate racist, sexist or otherwise toxic text with harmful intent. Many more use cases that could cause harm exist. Such applications to malicious use are discussed in detail in \citet{weidinger2021harms}.\\
    
    \bottomrule
    \caption{\textbf{\chinchilla model card.} We follow the framework presented in \citet{mitchell2019model}.
    }
    \label{tab:chinchilla-model-card}
\end{longtable}
\end{center}

\section{List of trained models}
In \autoref{tab:all_models} we list the model size and configuration of all models used in this study. 
Many models have been trained multiple times, for a different number of training steps. 
\clearpage
\footnotesize
\begin{longtable}[h!]
{r | rrrrr }
\toprule
Parameters (million) &  d\_model &  ffw\_size &  kv\_size &  n\_heads &  n\_layers \\
\midrule
    44 &                   512 &                          2048 &                     64 &                       8 &                        8 \\
    57 &                   576 &                          2304 &                     64 &                       9 &                        9 \\
    74 &                   640 &                          2560 &                     64 &                      10 &                       10 \\
    90 &                   640 &                          2560 &                     64 &                      10 &                       13 \\
   106 &                   640 &                          2560 &                     64 &                      10 &                       16 \\
   117 &                   768 &                          3072 &                     64 &                      12 &                       12 \\
   140 &                   768 &                          3072 &                     64 &                      12 &                       15 \\
   163 &                   768 &                          3072 &                     64 &                      12 &                       18 \\
   175 &                   896 &                          3584 &                     64 &                      14 &                       14 \\
   196 &                   896 &                          3584 &                     64 &                      14 &                       16 \\
   217 &                   896 &                          3584 &                     64 &                      14 &                       18 \\
   251 &                  1024 &                          4096 &                     64 &                      16 &                       16 \\
   278 &                  1024 &                          4096 &                     64 &                      16 &                       18 \\
   306 &                  1024 &                          4096 &                     64 &                      16 &                       20 \\
   425 &                  1280 &                          5120 &                    128 &                      10 &                       18 \\
   489 &                  1280 &                          5120 &                    128 &                      10 &                       21 \\
   509 &                  1408 &                          5632 &                    128 &                      11 &                       18 \\
   552 &                  1280 &                          5120 &                    128 &                      10 &                       24 \cr
   587 &                  1408 &                          5632 &                    128 &                      11 &                       21 \cr
   632 &                  1536 &                          6144 &                    128 &                      12 &                       19\cr
   664 &                  1408 &                          5632 &                    128 &                      11 &                       24\cr
   724 &                  1536 &                          6144 &                    128 &                      12 &                       22 \cr
   816 &                  1536 &                          6144 &                    128 &                      12 &                       25\cr
   893 &                  1792 &                          7168 &                    128 &                      14 &                       20\cr
  1,018 &                  1792 &                          7168 &                    128 &                      14 &                       23 \cr
  1,143 &                  1792 &                          7168 &                    128 &                      14 &                       26 \cr
  1,266 &                  2048 &                          8192 &                    128 &                      16 &                       22\cr
  1,424 &                  2176 &                          8704 &                    128 &                      17 &                       22 \cr 
  1,429 &                  2048 &                          8192 &                    128 &                      16 &                       25  \cr 
  1,593 &                  2048 &                          8192 &                    128 &                      16 &                       28  \cr
  1,609 &                  2176 &                          8704 &                    128 &                      17 &                       25 \cr 
  1,731 &                  2304 &                          9216 &                    128 &                      18 &                       24 \cr
  1,794 &                  2176 &                          8704 &                    128 &                      17 &                       28\cr 
  2,007 &                  2304 &                          9216 &                    128 &                      18 &                       28  \cr 
  2,283 &                  2304 &                          9216 &                    128 &                      18 &                       32  \cr  
  2,298 &                  2560 &                         10240 &                    128 &                      20 &                       26  \cr  
  2,639 &                  2560 &                         10240 &                    128 &                      20 &                       30 \cr  
  2,980 &                  2560 &                         10240 &                    128 &                      20 &                       34  \cr  
  3,530 &                  2688 &                         10752 &                    128 &                      22 &                       36  \cr 
  3,802 &                  2816 &                         11264 &                    128 &                      22 &                       36 \cr 
  4,084 &                  2944 &                         11776 &                    128 &                      22 &                       36 \cr 
  4,516 &                  3072 &                         12288 &                    128 &                      24 &                       36 \cr 
  6,796 &                  3584 &                         14336 &                    128 &                      28 &                       40 \cr 
  9,293 &                  4096 &                         16384 &                    128 &                      32 &                       42  \cr 
 11,452 &                  4352 &                         17408 &                    128 &                      32 &                       47  \cr  
 12,295 &                  4608 &                         18432 &                    128 &                      36 &                       44\cr  
 12,569 &                  4608 &                         18432 &                    128 &                      32 &                       47 \cr 
 13,735 &                  4864 &                         19456 &                    128 &                      32 &                       47 \cr 
 14,940 &                  4992 &                         19968 &                    128 &                      32 &                       49  \cr  
 16,183 &                  5120 &                         20480 &                    128 &                      40 &                       47 \cr  
\bottomrule
\caption{\textbf{All models.}
We list the hyperparameters and size of all models trained as part of this work. Many shown models have been trained with multiple learning rate schedules/number of training tokens.}
\label{tab:all_models}
\end{longtable}

\end{document}